\documentclass[lettersize,journal]{IEEEtran}

\usepackage{amsmath,amsfonts,amssymb}
\usepackage{algorithm}
\usepackage{algcompatible}
\usepackage{array}
\usepackage{booktabs}
\usepackage{cite}
\usepackage{diagbox}
\usepackage{enumitem}
\usepackage{float}
\usepackage{graphicx}
\usepackage{orcidlink}
\usepackage{textcomp}
\usepackage{stfloats}
\usepackage{url}
\usepackage{verbatim}
\usepackage{xcolor}
\usepackage{hyperref}

\let\oldSigma\Sigma
\renewcommand{\Sigma}{\mathrm{\oldSigma}}
\let\oldGamma\Gamma
\renewcommand{\Gamma}{\mathrm{\oldGamma}}

\usepackage{indentfirst}
\algnewcommand\INPUT{\item[\textbf{Input:}]}%
\algnewcommand\OUTPUT{\item[\textbf{Output:}]}%

\usepackage{booktabs}
\usepackage{multirow}
\usepackage{xcolor}

\begin{document}

\title{Hyperspectral Diffusion Equivariant Imaging (HyDiff-EI): A Self-supervised Framework for Hyperspectral Image Inpainting}

\author{Shuo~Li,~\IEEEmembership{Member,~IEEE},
        Mike~Davies,~\IEEEmembership{Fellow,~IEEE},
        and Mehrdad~Yaghoobi,~\IEEEmembership{Member,~IEEE}%
\thanks{Shuo Li is with the Institute for Digital Communications, School of Engineering, University of Edinburgh. E-mail: sli16@ed.ac.uk.}%
\thanks{Mike Davies is with the School of Engineering, University of Edinburgh. E-mail: mike.davies@ed.ac.uk.}%
\thanks{Mehrdad Yaghoobi is with the School of Engineering, University of Edinburgh. E-mail: m.yaghoobi-vaighan@ed.ac.uk.}}

\markboth{IEEE Transactions on Computational Imaging}%
{Li \MakeLowercase{\textit{et al.}}: HyDiff-EI for Hyperspectral Image Inpainting}

\maketitle

\begin{abstract}

A novel Hyperspectral diffusion Equivariant Imaging (HyDiff-EI) framework for solving the hyperspectral image (HSI) inpainting problem has been presented here. Unlike conventional diffusion-based methods that rely on large-scale pretraining, HyDiff-EI is a test-time optimization framework that learns directly from a single corrupted HSI acquisition. This makes it flexible for different sensor configurations and particularly well-suited for practical remote sensing scenarios where large annotated hyperspectral datasets are limited. To address the ill-posed nature of unsupervised inpainting, we embed equivariant consistency constraints within the diffusion process. By leveraging the inherent geometric symmetries and intrinsic characteristics of HSIs, HyDiff-EI bridges the gap between generative diffusion modeling and self-consistent physical priors. We empirically show that coupling diffusion modeling with equivariant priors substantially enhances noise robustness and generalizability. Extensive experiments on real-world datasets including Chikusei, Botswana, and EMIT demonstrate that HyDiff-EI offers remarkable inpainting quality over existing self-supervised and diffusion-based algorithms in both noiseless and noisy cases. 
\end{abstract}


\section{Introduction}
\noindent
Hyperspectral imaging captures rich spatial-spectral information across a wide range of wavelengths. However, due to various factors such as sensor failures or data transmission errors, certain areas of the HSIs acquired in practice are incomplete or corrupted. HSI inpainting refers to the process of filling in missing pixels/bands in incomplete observations, which plays a crucial role in ensuring data completeness, facilitating accurate analysis and enhancing visualization. {As a solution, self-supervised HSI inpainting methods \cite{DHP, pnp-dip, niresi2023robust, li2025self}} offer several advantages over heavily trained methods \cite{HSI-IPNet,HSI-LPIN}, including: (1) Flexibility: it becomes less required to train or re-train different models on different datasets; (2) Generalization ability: it generalizes well to unseen data. These advantages make it especially suitable for processing HSIs as gathering HS data and training with large-scale models could be expensive. 

\subsection{Related Works}
\noindent
The diffusion model belongs to a broader class of deep generative models which has been gaining increasing popularity in the field of deep learning and computer vision. Benefiting from its interpretable structures by design, diffusion models are able to capture complex data distributions and generate high-quality and diverse samples \cite{DDPM,diffusion_beats_gans, score_based_diffusion}. The strong learning capability of the diffusion model makes it a potential candidate for solving the image inpainting task. For example, in \cite{ILVR}, authors incorporate the reference image into the DDPM reverse process to guide the image generation, making "conditional" sampling possible. As an extension, the authors in \cite{DDPM_repaint} considered the reference image to be incomplete, i.e., there are missing regions to be filled in and proposed an inpainting algorithm aided by some time-travel tricks to improve the performance. In \cite{DDRM}, the reverse diffusion process is reformulated to address a broader class of inverse problems, marking one of the first mathematically grounded uses of diffusion models as inverse solvers. In \cite{DDNM}, the authors borrowed the concept from linear algebra and suggested a Range-Null Space Decomposition (RND) at each diffusion step, which further concluded \cite{DDRM} as a special case. Later works \cite{Diffusion_PnP,rout2023solving,wang2024dmplug, rout2024beyond} have further demonstrated the effectiveness of using diffusion models for RGB image inpainting. For a comprehensive overview of these advancements, readers are referred to the review in \cite{daras2024survey}. More recently, diffusion models have also been adopted within the remote sensing community and successfully applied to various HSI inverse tasks, including denoising, restoration, and super-resolution \cite{xiao2023ediffsr, meng2024conditional_SR, li2024latent_Restoration, pang2024hir}, as extensively reviewed in \cite{hu2025recent}. Among these works, \cite{pang2024hir} proposed to estimate the reduced HS image with the pre-trained diffusion models and bring it back to the full spectral range for reconstruction, achieving promising performance across multiple HSI restoration tasks.

However, the success of these aforementioned methods relies heavily on the expressive power of the pre-trained diffusion models. Applying these models to HSI inverse problems typically requires training on large-scale hyperspectral datasets \cite{DDPM}, which can be computationally demanding and impractical. In this work, we aim to leverage the potential of diffusion models on the HSI inpainting task and propose a new framework that does not impose any burden on network training. To this end, we develop a self-supervised diffusion-based HSI inpainting algorithm, HyDiff-EI, based on the recently popular equivariant imaging (EI) priors \cite{chen2021equivariant, tachella2024unsure, Hyper-EI, scanvic2025scale}. Unlike existing DHP-based methods \cite{DHP, niresi2023robust, pnp-dip, miao2023dds2m,li2025self}, HyDiff-EI leverages known physical or geometric symmetries as priors rather than relying solely on network architecture. Moreover, it operates as a self-supervised diffusion framework and does not require pre-trained diffusion models \cite{pang2024hir}, whether trained on extensive remote sensing or RGB datasets. This endows the method with extra flexibility: it is sensor-agnostic that can naturally accommodate HSIs with different numbers of spectral bands. While pre-trained models are fundamentally constrained by their specific training distributions and noise levels, HyDiff-EI seamlessly adapts to unseen sensor configurations, making it highly suitable for real-world deployments.

\subsection{Contribution}
\noindent This paper presents a powerful HS inpainting algorithm that dynamically combines self-supervised learning with recent diffusion models. Our main contributions are:
\begin{itemize}
\item[\textbullet] To the best of our knowledge, this is the first work to incorporate Equivariant Imaging (EI) \cite{chen2021equivariant} within the discrete-time diffusion framework for hyperspectral image inpainting. Unlike methods relying on unstructured unsupervised priors (such as those based on DHP), our approach learns directly from geometric invariance structures.\footnote{A concurrent work \cite{levac2025normalization} (not yet published) introduces a continuous-time score-based formulation; however, its scalability when applied to hyperspectral inpainting remains to be investigated.}

\item[\textbullet] Extensive ablation studies demonstrate that integrating diffusion with EI consistently improves inpainting performance over conventional EI, achieving substantial gains in the noiseless setting and maintaining robust performance in the noisy settings.

\item[\textbullet] The proposed method achieves state-of-the-art performance on real-world HSI inpainting tasks, demonstrating significant improvements over existing deep generative prior–based approaches \cite{niresi2023robust, miao2023dds2m, pang2024hir}.
\end{itemize}
The rest of this paper is organized as follows: {Section \ref{section_background}
presents the preliminaries of EI and DDPM, and introduces our proposed HyDiff-EI algorithm.}
Section \ref{experiments} discusses the implementation details, followed by the experiments on three different HSI datasets. Section \ref{section_ablations} discusses the ablation studies. Finally, Section \ref{conclusion} concludes this paper.
\section{Background} \label{section_background}
\subsection{Transformation Invariance in HSI Inpainting}
\noindent The HSI inpainting task can be formulated as the recovery of the clean image $\boldsymbol{x}$ from the corrupted measurement $\boldsymbol{y}$, in the presence of additive noise $\boldsymbol{n}$ and a binary masking matrix $\rm M$:
\begin{equation} \label{Formulation_HSI_Inpainting}
 \boldsymbol{y} = {\rm M} \boldsymbol{x} +\boldsymbol{n}
\end{equation}

Due to the ill-posed nature of the problem, it is essential to incorporate either hand-crafted or learned priors of the clean HS image $\boldsymbol{x}$ to regularize the reconstruction process. The equivariant imaging (EI) method exploits the invariance of typical image distributions to transformations such as rotations and translations, making it possible to learn directly from the measurement data itself. In the context of HSI inpainting, transformations applied to the input corrupted HS image should be reflected in a predictable way in the reconstructed HS image. e.g., A shift or rotation in the observed image should result in the same shift or rotation in the recovered clean image. Preserving such equivariant properties is crucial for HSI inpainting algorithms to accurately model the underlying acquisition and degradation process.

Following the definitions in \cite{chen2021equivariant}, we denote the group transformation action as $\mathcal{G}=\{\boldsymbol{g}_1,\boldsymbol{g}_2,\ldots,\boldsymbol{g}_{N_g}\}$, and a set of signals as $\mathcal{X} \subseteq \mathbb{R}^{q}$. Then, for unitary group actions $T_{g}$, and for all the signals $\boldsymbol{x} \in \mathcal{X}$, the group invariance property assumes:
\begin{equation} \label{group_invariance_assump}
 T_{g} \boldsymbol{x} \in \mathcal{X}
\end{equation}
\noindent it directly follows that:
\begin{equation} \label{EI virtual operator}
 \boldsymbol{y} = {\rm M}(\boldsymbol{x}) = {\rm M}(T_{g}T_{g}^{-1} \boldsymbol{x}) = {\rm M}_g T_{g}^{-1}(\boldsymbol{x})
\end{equation}

One can see that group transformation is capable of exploiting new sets of virtual measurement operators ${\rm M}_g$ that possibly span different null-spaces, thus providing extra information for recovering the clean image $\boldsymbol{x}$. Combining \eqref{group_invariance_assump} with a noiseless measurement matrix ${\rm M}$ and the reconstruction model $f_{\theta}$, we obtain the following transformation-equivariant constraint:
\begin{equation} \label{EI constraint}
 f_{\theta}({\rm M}T_{g} \boldsymbol{x}) = T_{g} f_{\theta}({\rm M} \boldsymbol{x})
\end{equation}
\noindent that is, the composition of $f_{\theta}\circ \rm M$ should be equivariant to applied transformations. In practice, however, the ground-truth clean image $\boldsymbol{x}$ is unavailable. Therefore, the constraint in \eqref{EI constraint} is implemented empirically as:
\begin{equation} \label{EI constraint emprical}
 f_{\theta}(\mathrm{M} (T_{g} f_{\theta}(\boldsymbol{y}))) = T_{g} f_{\theta}(\boldsymbol{y})
\end{equation}
\noindent where the right-hand-side represents the transformed network output $\tilde{\boldsymbol{x}}^{'} = T_{g} f_{\theta}(\boldsymbol{y})$. On the left-hand-side, we re-inject the masked version of this transformed prediction $\mathrm{M} \tilde{\boldsymbol{x}}^{'}$ back into the same EI net $f_{\theta}(\cdot)$ to produce a secondary output $\tilde{\boldsymbol{x}}^{''}$, in which way $\tilde{\boldsymbol{x}}^{''}$ should agree with $\tilde{\boldsymbol{x}}^{'}$ due to the geometric symmetries.
\begin{figure*}[h]
\centering
  \scalebox{0.9}{
  \includegraphics[width=0.82\textwidth,height=0.3\textwidth,]{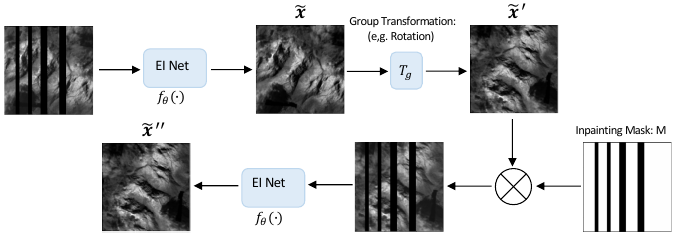}
 }
  \caption{EI Workflow. $\tilde{\boldsymbol{x}}^{'}$ and $\tilde{\boldsymbol{x}}^{''}$ correspond to the right-hand side and left-hand side of Eq. \eqref{EI constraint emprical}, respectively.}
  \label{Enforcing_EI_Loss}  
  \vspace{-0.5cm}
\end{figure*} 

\noindent \textbf{Enforcing EI Constraint through Training.} Figure \ref{Enforcing_EI_Loss} illustrates the workflow of the EI framework, where $\tilde{\boldsymbol{x}}$ denotes the estimated clean image produced by the EI network $f_{\theta}$. The transformation $T_{g}$ is applied to obtain $\tilde{\boldsymbol{x}}^{'}$, which is then passed through the composition operator $f_{\theta}\circ \rm M$ to generate $\tilde{\boldsymbol{x}}^{''}$, a re-estimate of $\tilde{\boldsymbol{x}}^{'}$. The EI constraint in Eq.~\eqref{EI constraint emprical} is enforced during training by penalizing the discrepancy between $\tilde{\boldsymbol{x}}^{'}$ and $\tilde{\boldsymbol{x}}^{''}$, encouraging equivariant consistency in the reconstruction process.

\subsection{Diffusion Models}
\noindent The denoising diffusion probabilistic model (DDPM) \cite{DDPM} defines two processes: a forward process that gradually adds Gaussian noise to data up to $T$ time steps, and a backward process that iteratively searches for the good estimation of the clean data, hence denoising. Denote $\boldsymbol{x}_0$ as the clean data, the forward/diffusion process can be written as a Markov chain with respect to some predefined variance schedule $\beta_1,...,\beta_t$: 
\begin{equation}
\begin{aligned} \label{forward_pass}
q(\boldsymbol{x}_t|\boldsymbol{x}_{t-1}) &:= \mathcal{N}(\boldsymbol{x}_{t};  \sqrt{1-\beta_t}\boldsymbol{x}_{t-1}, \beta_t {\rm I}) \\
q(\boldsymbol{x}_{1:T}|\boldsymbol{x}_0)&:= \prod_{t=1}^T q(\boldsymbol{x}_t|\boldsymbol{x}_{t-1})
\end{aligned}
\end{equation}

Similarly, the backward/denoising process is defined as the joint probability $p(\boldsymbol{x}_{0:T})$. Starting from the random noise $\boldsymbol{x}_T$, this process aims at generating the previous state $\boldsymbol{x}_{t-1}$ from the current state $\boldsymbol{x}_t$:
\begin{equation}
\begin{aligned} \label{backward_pass}
p_{\theta}(\boldsymbol{x}_{t-1}|\boldsymbol{x}_t) &:= \mathcal{N}(\boldsymbol{x}_{t-1};  \boldsymbol{\mu}_{\theta}(\boldsymbol{x}_t,t), \boldsymbol{\Sigma}_{\theta}(\boldsymbol{x}_t,t)), \\
p_{\theta}(\boldsymbol{x}_{0:T})&:= p_{\theta}(\boldsymbol{x}_T)\prod_{t=1}^T p_{\theta}(\boldsymbol{x}_{t-1}|\boldsymbol{x}_{t})
\end{aligned}
\end{equation}
where $\theta$ stands for the parameters of the trained DDPM neural network: $\epsilon_{\theta}(\boldsymbol{x}_t,t)$, $\boldsymbol{\mu}_{\theta}$ and $\boldsymbol{\Sigma}_{\theta}$ are the predicted posterior mean and variance, respectively. {By optimizing the variational bound on the negative log-likelihood}, $p_{\theta}(\boldsymbol{x}_{t-1}|\boldsymbol{x}_t)$ can be calculated from $q(\boldsymbol{x}_t|\boldsymbol{x}_{t-1})$ (please see the derivation in \cite{DDPM}), the resulting $\boldsymbol{\mu}_{\theta}$ and $\boldsymbol{\Sigma}_{\theta}$ take the form: \\
\begin{equation}
\begin{aligned} \label{DDPM_preliminary}
\boldsymbol{\mu}_{\theta}(\boldsymbol{x}_t,t)&:= \frac{\sqrt{\bar{\alpha}_{t-1}}\beta_t}{1-\bar{\alpha}_{t}} \boldsymbol{x}_0 + \frac{\sqrt{\alpha}_{t}(1-\bar{\alpha}_{t-1})}{1-\bar{\alpha}_{t}} \boldsymbol{x}_t \\
&:= \frac{1}{\sqrt{\alpha_t}}(\boldsymbol{x}_t -\frac{1-\alpha_{t}}{\sqrt{1-\bar{\alpha}_{t}}}\epsilon_{\theta}(\boldsymbol{x}_t,t) )\\
\boldsymbol{\Sigma}_{\theta}(t) &:= \frac{1-\bar{\alpha}_{t-1}}{1-\bar{\alpha}_{t}} \beta_t I, \text{where}
\begin{cases} \alpha_t:=1-\beta_t& \\\bar{\alpha}_t:=\prod_{i=1}^t\alpha_i& \end{cases}
\end{aligned}
\end{equation}

The clean data $\boldsymbol{x}_0$ can thus be reconstructed by iteratively drawing samples from $p_{\theta}(\boldsymbol{x}_{t-1}|\boldsymbol{x}_t)$ using Eq. \eqref{DDPM_preliminary}, from time step $t=T$ down to $t=0$.

\subsection{Proposed Method}
\noindent Conventional DDPM requires training the model on large datasets in order to achieve state-of-the-art performance \cite{DDPM,DDIM,DDRM,DDPM_repaint}, which can be cumbersome for data-hungry HSI applications. We propose to use a self-supervised equivariant imaging neural network {$f_{\theta}^{(t)}(\boldsymbol{y})$} to replace the DDPM net: $\epsilon_{\theta}(\boldsymbol{x}_t,t)$. Different from DDPM where the network is trained as a denoiser, we train the EI net: {$f_{\theta}^{(t)}(\boldsymbol{y})$} to directly estimate clean data: $\boldsymbol{x}_{0|t}$ at time step $t$. This motivates us to re-design the DDPM framework to accommodate the EI training process. Specifically, we modify the EI loss function as follows (please refer to the supplementary material for detailed derivation):
\begin{equation}
\begin{aligned} \label{EI_loss}
  L^{(t)} = \left \Vert \boldsymbol{x}_t-\sqrt{1-\bar{\alpha}_t} \epsilon_0 - \sqrt{\bar{\alpha}_t}f_{\theta}^{(t)}(\boldsymbol{y}) 
\right \Vert^2_2 \\ + \alpha_{EI} \Vert\tilde{\boldsymbol{x}_t}^{'}- \tilde{\boldsymbol{x}_t}^{''}\Vert^2_2
\end{aligned}
\end{equation} 
\noindent where $\epsilon_0 \sim \mathcal{N}(0,{\rm I})$ is the noise added for each time step $t$ in the forward/diffusion process. The first term on the right-hand side of Eq.~\eqref{EI_loss} corresponds to the data-fidelity term, while the second term, weighted by $\alpha_{EI}$, involves both $\tilde{\boldsymbol{x}_t}^{'}$ and $\tilde{\boldsymbol{x}_t}^{''}$, which are computed at each time step $t$ following the procedure illustrated in Fig.~\ref{Enforcing_EI_Loss}.
Once the network parameters $\theta$ are updated, we can substitute $\boldsymbol{x}_0$ with {$f_{\theta}^{(t)}(\boldsymbol{y})$ in the posterior mean estimator \eqref{DDPM_preliminary}.} Note that it is also possible to learn the variance, however, {our experiments show that} it might lead to unstable training and degraded sampling quality, so we keep the variance fixed as suggested in \cite{DDPM}. We adopt the structures in \cite{DDNM} as the backbone, which is a modified version of DDPM applied on various image inverse tasks. Denote the inpainting mask as $\rm M$ and its pseudoinverse as $\rm M^{\dag}$, the clean data $\boldsymbol{x}_{0|t}$ is then re-estimated as: 
\begin{equation}
\begin{aligned} \label{re_estimation}
 \tilde{\boldsymbol{x}}_{0|t} = {\rm M^{\dag}} \boldsymbol{y} + ({\rm I}-{\rm M \rm M^{\dag}}) f_{\theta}^{(t)}(\boldsymbol{y})
\end{aligned}
\end{equation}

Step \eqref{re_estimation} is the core idea of DDNM. More specifically, it decomposes a sample into range-space and null-space parts and iteratively refines only the null-space contents during the reverse diffusion sampling. In the HSI inpainting problem, we propose to refine the null-space part using the contents generated by EI. The proposed HyDiff-EI algorithm in the noiseless case is presented in Algorithm \ref{HyDiff algorithm}.
\begin{algorithm}[H]
    \caption{(HyDiff-EI) Algorithm} \label{HyDiff algorithm}
  \begin{algorithmic}
\REQUIRE degradation matrix: $\rm M$, corrupted HSI: $\boldsymbol{y}$, diffusion steps: $T$, {EI net: $f_{\theta}^{(t)}(\cdot)$.}
    
    \OUTPUT reconstructed HSI image $\boldsymbol{x}_0$.

   \STATE \textbf{Initialization} EI network parameters: $\theta$, EI Input: $\boldsymbol{y}$, keep fixed.
    \STATE 1. diffuse $\boldsymbol{y}$ up to $T$ step to get initial estimate: $\boldsymbol{x}_T$, using \eqref{forward_pass}.
   \FOR{$t = T,...,1 $}:
        
      \STATE 2. update EI network parameters $\theta$ by minimizing \eqref{EI_loss}. 
      \STATE 3. re-estimate $\tilde{\boldsymbol{x}}_{0|t}$: \par
          \hskip\algorithmicindent  \hskip\algorithmicindent 
          \hskip\algorithmicindent 
          $\tilde{\boldsymbol{x}}_{0|t} = {\rm M^{\dag}} \boldsymbol{y} + ({\rm I}-{\rm M \rm M^{\dag}}) f_{\theta}^{(t)}(\boldsymbol{y})$
        
      \STATE 4. sample $\boldsymbol{x}_{t-1}$ from $p(\boldsymbol{x}_{t-1}|\boldsymbol{x}_t,\tilde{\boldsymbol{x}}_{0|t})$:\par
      
           \hskip\algorithmicindent  \hskip\algorithmicindent  \hskip\algorithmicindent  \hskip\algorithmicindent  \hskip\algorithmicindent 
           $\boldsymbol{\epsilon} \sim \mathcal{N}(0,\,{\rm I})$  \par
           
             \hskip\algorithmicindent  \hskip\algorithmicindent  \hskip\algorithmicindent  \hskip\algorithmicindent  \hskip\algorithmicindent 
           $\sigma_t^2=\frac{1-\bar{\alpha}_{t-1}}{1-\bar{\alpha}_{t}} \beta_t$  \par
        
 \hskip\algorithmicindent 
         $\boldsymbol{x}_{t-1} = \frac{\sqrt{\bar{\boldsymbol{\alpha}}_{t-1}}\boldsymbol{\beta}_t}{1-\bar{\boldsymbol{\alpha}}_{t}} \tilde{\boldsymbol{x}}_{0|t} + \frac{\sqrt{\boldsymbol{\alpha}}_{t}(1-\bar{\boldsymbol{\alpha}}_{t-1})}{1-\bar{\boldsymbol{\alpha}}_{t}} \boldsymbol{x}_t + \sigma_t\boldsymbol{\epsilon} $
    \ENDFOR
\STATE return $\boldsymbol{x}_0$
  \end{algorithmic}
\end{algorithm} 

\subsection{Extension to Noisy HSI Inpainting}
\noindent We propose a variant of the HyDiff-EI algorithm in noisy case (Algorithm \ref{HyDiff-EI-noisy algorithm}). Inspired by DDNM \cite{DDNM}, we modify the re-estimation step \eqref{re_estimation} in the noisy case as:
\begin{equation}
\begin{aligned} \label{re_estimation_noisy_case}
 \tilde{\boldsymbol{x}}_{0|t} = f_{\theta}^{(t)}(\boldsymbol{y}) - {\color{red}\Sigma_t} {\rm M^{\dag}}({\rm M} f_{\theta}^{(t)}(\boldsymbol{y}) -\boldsymbol{y})
\end{aligned}
\end{equation}
\noindent {where ${f_{\theta}^{(t)}(\boldsymbol{y})}$ is the EI network output, and {$\Sigma_t = {\rm U} \operatorname{diag} \{ \lambda_{t1},\lambda_{t2}...\lambda_{tq} \} \rm V^T$} is a scaling matrix \footnote{{In this formulation, $\rm U$ and $\rm V^T$ are derived from the singular value decomposition (SVD) of the inpainting mask $\rm M$, Notation $\lambda_{ti}$ is adopted instead of $\sigma_{ti}$ to avoid confusion with $\sigma_{t}$ used in Eq. \eqref{Sigma_t_main_body} and \eqref{introduced_noise}} }that scales the range-space correction term based on the estimated HSI at each time step $t$.}

\begin{algorithm}[t]
    \caption{(HyDiff-EI) Algorithm in Noisy Case} \label{HyDiff-EI-noisy algorithm}
  \begin{algorithmic}
\REQUIRE degradation matrix: $\rm M$, corrupted HSI: $\boldsymbol{y}$, noise strength: $\sigma_y$, diffusion steps: $T$, {EI net: $f_{\theta}^{(t)}(\cdot)$.}
    
    \OUTPUT reconstructed HSI image $\boldsymbol{x}_0$.

   \STATE \textbf{Initialization} EI network parameters: $\theta$, EI Input: $\boldsymbol{y}$, keep fixed.
    \STATE 1. diffuse $\boldsymbol{y}$ up to $T$ step to get initial estimate: $\boldsymbol{x}_T$, using \eqref{forward_pass}.
   \FOR{$t = T,...,1 $}:
        
      \STATE 2. update EI network parameters $\theta$ by minimizing \eqref{EI_loss}. 

      \STATE 3. re-estimate $\tilde{\boldsymbol{x}}_{0|t}$: \par
          \hskip\algorithmicindent  \hskip\algorithmicindent 
          \hskip\algorithmicindent 
          $\tilde{\boldsymbol{x}}_{0|t} = {f_{\theta}^{(t)}(\boldsymbol{y})} - {\color{red}\Sigma_t} {\rm M}^{\dag}({\rm M}{f_{\theta}^{(t)}(\boldsymbol{y})} -\boldsymbol{y})$
        
      \STATE 4. sample $\boldsymbol{x}_{t-1}$ from $p(\boldsymbol{x}_{t-1}|\boldsymbol{x}_t,\tilde{\boldsymbol{x}}_{0|t})$:\par
      
           \hskip\algorithmicindent  \hskip\algorithmicindent  \hskip\algorithmicindent  \hskip\algorithmicindent  \hskip\algorithmicindent 
           $\boldsymbol{\epsilon} \sim \mathcal{N}(0,\,{\rm I})$  \par
           
             \hskip\algorithmicindent  \hskip\algorithmicindent  \hskip\algorithmicindent  \hskip\algorithmicindent  \hskip\algorithmicindent 
           $\sigma_t^2=\frac{1-\bar{\alpha}_{t-1}}{1-\bar{\alpha}_{t}} \beta_t$  \par
        
 \hskip\algorithmicindent 
         $\boldsymbol{x}_{t-1} = \frac{\sqrt{\bar{\boldsymbol{\alpha}}_{t-1}}\boldsymbol{\beta}_t}{1-\bar{\boldsymbol{\alpha}}_{t}} \tilde{\boldsymbol{x}}_{0|t} + \frac{\sqrt{\boldsymbol{\alpha}}_{t}(1-\bar{\boldsymbol{\alpha}}_{t-1})}{1-\bar{\boldsymbol{\alpha}}_{t}} \boldsymbol{x}_t  +\boldsymbol{\epsilon}_{\mathrm{con}} $
    \ENDFOR
\STATE return $\boldsymbol{x}_0$
  \end{algorithmic}
\end{algorithm}
 Intuitively, when the observed image is highly noisy, there exists a stage during the algorithm’s progression where the range-space information becomes unreliable, since the model has already developed a strong belief of what a clean image should look like, the noisy observation contributes little useful information. Consequently, less weight should be given to the observation. When $\Sigma_t$ is set to the identity matrix $\rm I$, equation \eqref{re_estimation_noisy_case} reduces to \eqref{re_estimation}. In the noiseless HSI inpainting scenario, $\Sigma_t$ is the identity matrix to preserve the range-space information. However, when the observation itself is noisy, {$\Sigma_t$ should be chosen such that the scaled range-space and null-space contents remains consistent.} The update of $\Sigma_t$ follows the formulation in DDNM \cite{DDNM} and is given as:
\begin{equation} \label{Sigma_t_main_body}
 \lambda_{ti} = \left\{
\begin{array}{l}
1 \qquad\qquad\qquad \sigma_t \ge \frac{(\frac{\sqrt{\bar{\alpha}_{t-1}}\beta_t}{1-\bar{\alpha}_{t}}\sigma_y}{s_i} \\
\frac{\sigma_ts_i}{\frac{\sqrt{\bar{\alpha}_{t-1}}\beta_t}{1-\bar{\alpha}_{t}}\sigma_y} \qquad \sigma_t < \frac{(\frac{\sqrt{\bar{\alpha}_{t-1}}\beta_t}{1-\bar{\alpha}_{t}}\sigma_y}{s_i} \\
1  \qquad \qquad \qquad s_i = 0 \\
\end{array}
\right.
\end{equation}

\noindent where $s_i$ denotes the $i$-th singular value of the mask $\mathrm{M} = \mathrm{U} \operatorname{diag}\{s_1, s_2, \dots, s_q\} \mathrm{V}^T$. \footnote{Note that for the inpainting task, $\rm M$ is row-orthonormal, so $s_i$ can be simplified to 1 or 0. Nevertheless, we keep $s_i$ to allow generalization to other operators.} $\sigma_t$ and $\sigma_y$ are the noise strength of the current estimate $\boldsymbol{x}_{t-1}$ and the noise strength of the observation $\boldsymbol{y}$, respectively. Besides, we also introduce an extra noise term $\boldsymbol{\epsilon}_{\mathrm{con}}$ into the output step \eqref{DDPM_preliminary}:
\begin{equation} \label{introduced_noise}
\begin{aligned}
&\qquad \quad \boldsymbol{\epsilon}_{\mathrm{con}} \sim \mathcal{N}(\boldsymbol{0},\boldsymbol{\Gamma}_t),\\
&\boldsymbol{\Gamma}_t = \operatorname{diag}\{\Gamma_{t1},\Gamma_{t2},\ldots,\Gamma_{tq}\}, \\
&\Gamma_{ti} = \begin{cases}
 \sigma_t^2 - \frac{(\frac{\sqrt{\bar{\alpha}_{t-1}}\beta_t}{1-\bar{\alpha}_{t}})^2 
\sigma_y^2 \lambda_{ti}^2}{s_i^2}   & s_i \neq 0 \\
 \sigma_t^2  & s_i = 0 \\
\end{cases}
\end{aligned}
\end{equation}
\noindent so that \eqref{DDPM_preliminary} becomes: 
\begin{equation} \label{modified_output}
\begin{aligned}
  & \qquad \qquad \qquad \qquad \quad \boldsymbol{\epsilon} \sim \mathcal{N}(0,\,{\rm I}) \\
& \boldsymbol{x}_{t-1} = \frac{\sqrt{\bar{\boldsymbol{\alpha}}_{t-1}}\boldsymbol{\beta}_t}{1-\bar{\boldsymbol{\alpha}}_{t}} \tilde{\boldsymbol{x}}_{0|t} + \frac{\sqrt{\boldsymbol{\alpha}}_{t}(1-\bar{\boldsymbol{\alpha}}_{t-1})}{1-\bar{\boldsymbol{\alpha}}_{t}} \boldsymbol{x}_t + \boldsymbol{\epsilon}_{\mathrm{con}}
\end{aligned}       
\end{equation}
\begin{figure*}[t]
\vspace{-0.2cm}
\centering
  \includegraphics[width=1.02\textwidth, height=0.46\textwidth]{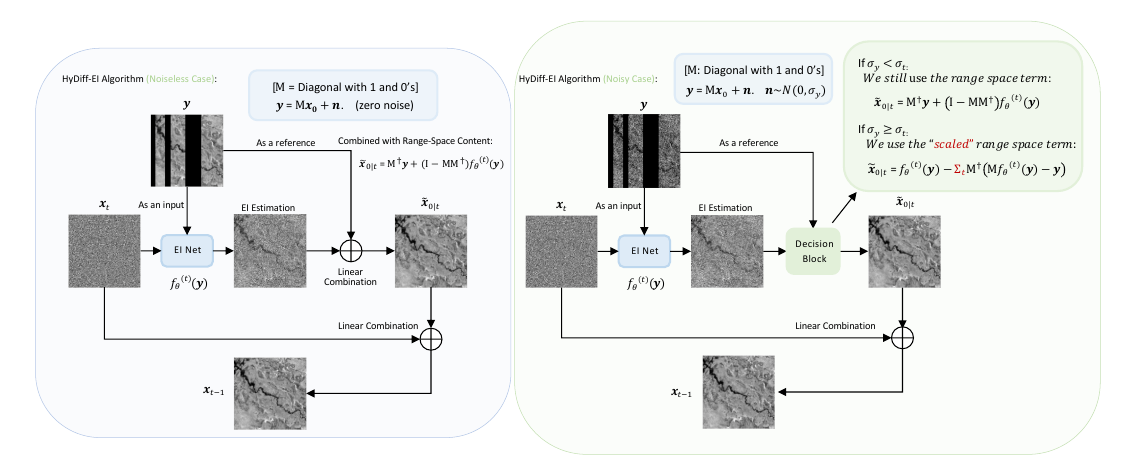}
  \caption{Flow Chart of the Proposed HyDiff-EI Algorithm. From Left to Right: Noiseless Case and Noisy Case.}
  \label{Network_Structure}  
  \vspace{-0.5cm} % <--- Adjust this negative value to shrink the bottom margin
\end{figure*}

The reason for making this adaption is that with the presence of noise in the observed image $\boldsymbol{y}$, the total noise variance in the estimated image $\boldsymbol{x}_{t-1}$ at each single update should conform to the noise level defined in the forward/diffusion step \eqref{forward_pass}. We place the derivation of equation \eqref{Sigma_t_main_body} and \eqref{introduced_noise} in the supplementary material for completeness. 
\section{Experiments}\label{experiments}
\subsection{Datasets and Implementation Details}
\noindent We initially evaluate the proposed HyDiff-EI Algorithm on two publicly available hyperspectral datasets: (1) The Chikusei airborne hyperspectral dataset  \cite{Chikusei} taken by Headwall Hyperspec-VNIR-C imaging sensor. The test image has 128 channels in the spectral range from 363 to 1018 nm{, and} (2) The Botswana dataset, collected by the NASA EO-1 satellite \cite{Botswana_dataset}, comprising 145 spectral bands covering the range of 400–2500 nm. {We then generate a dataset of real-time data} acquired by the NASA Earth Surface Mineral Dust Source Investigation (EMIT) mission \cite{EMIT_Mission}, launched in 2022. The test image includes 285 spectral bands across the 380–2500 nm range. For consistency in evaluation, all hyperspectral images are cropped to a spatial size of 144 × 144 pixels. For the implementation of the neural network ${f_{\theta}^{(t)}(\boldsymbol{y})}$, we adopt the same Skip-Net architecture as in DHP \cite{DHP} and use Adam optimizer with a learning rate of 0.01. For the choice of group transformations in EI, we use random $90 ^{\circ}$ rotations as the group transformation $T_{g}$ with a group size $N_g=7$. For the strength of the equivariant constraint term, we do not fine-tune the weight $\alpha_{EI}$ but set it to be 1 across all the tests. On the implementation of the diffusion process, we use DDNM \cite{DDNM} as the backbone, with T = 1000 timesteps, noise scaling factors $\boldsymbol{\beta}_{start}$ = 0.0001, $\boldsymbol{\beta}_{end}$ = 0.02. Similar to \cite{DDRM} and \cite{DDNM}, we treat the corrupted input image/reference image $\boldsymbol{y}$ differently depending on its noise level. Intuitively, if the noise level of the observation $\boldsymbol{y}$ at a particular time step $t$ is larger than the noise level of the current state $\boldsymbol{x}_t$, then less information in $\boldsymbol{y}$ should be considered. To construct the noisy samples, we firstly corrupt them with the Gaussian noise with noise level $\sigma_y =(0.1,0.2,0.3)$, and then, generate different shapes of masks to mask out entire spectral bands. All algorithms are trained and tested on an NVIDIA Tesla V100 GPU.

\subsection{Results and Comparison}
\noindent We evaluate the proposed HyDiff-EI against representative self-supervised hyperspectral image inpainting methods, including Deep Hyperspectral Prior (DHP) \cite{DHP}, R-DLRHyIn \cite{niresi2023robust}, DDS2M \cite{miao2023dds2m}, HIR-Diff \cite{pang2024hir}, and the recent state-of-the-art method SHARE \cite{xie2026share}. To ensure the fairness of the comparison, DHP, R-DLRHyIn, and HyDiff-EI are implemented using the same 2D Skip-Net backbone, namely a U-Net with skip connections. For DDS2M and SHARE, we retain the network architectures specified in their original implementations. The optimization iterations, learning rates, diffusion-related configurations for DDS2M, HIR-Diff, and HyDiff-EI, and regularization weights for DHP and R-DLRHyIn are selected according to the recommendations of the respective papers. Unless otherwise specified, all methods use the same input data, degradation model, mask, and evaluation protocol. To account for the stochasticity introduced by random network initialization and diffusion sampling, each method is independently evaluated 3 times for every test sample, and the reported results are averaged over these runs.

\begin{table*}[!h]
    \centering
    \caption{\small
    Comparison between our proposed algorithm and other learning-based
    inpainting algorithms on the Chikusei and Botswana datasets in the
    noiseless case. The mean and standard deviation over 20 samples are reported.
    }
    \label{Table_HSI_Clean}

    \vspace{-0.3cm}
    \setlength{\tabcolsep}{4pt}
    \renewcommand{\arraystretch}{1.15}

    \resizebox{\linewidth}{!}{%
    \begin{tabular}{crrrrrrr}
        \toprule
        Methods
        & Input
        & DHP \cite{DHP}
        & R-DLRHyIn \cite{niresi2023robust}
        & HIR-Diff \cite{pang2024hir}
        & SHARE \cite{xie2026share}
        & DDS2M \cite{miao2023dds2m}
        & HyDiff-EI (\textcolor{red}{proposed}) \\
        \midrule

        MPSNR $\uparrow$
        & 23.401
        & $29.695\,(\pm 0.53)$
        & $33.307\,(\pm 0.51)$
        & $33.862\,(\pm 0.39)$
        & $34.089\,(\pm 1.03)$
        & $34.344\,(\pm 0.60)$
        & $\boldsymbol{35.791}\,(\pm 0.45)$ \\

        MSSIM $\uparrow$
        & 0.205
        & $0.860\,(\pm 0.01)$
        & $0.906\,(\pm 0.01)$
        & $0.913\,(\pm 0.01)$
        & $0.918\,(\pm 0.01)$
        & $0.921\,(\pm 0.01)$
        & $\boldsymbol{0.940}\,(\pm 0.01)$ \\

        SAM ($^\circ$) $\downarrow$
        & --
        & $3.139\,(\pm 0.91)$
        & $2.790\,(\pm 0.55)$
        & $3.270\,(\pm 0.55)$
        & $3.041\,(\pm 0.43)$
        & $2.264\,(\pm 0.45)$
        & $\boldsymbol{1.224}\,(\pm 0.47)$ \\

        SCC $\uparrow$
        & --
        & $0.985\,(\pm 0.01)$
        & $0.987\,(\pm 0.01)$
        & $0.984\,(\pm 0.01)$
        & $0.987\,(\pm 0.01)$
        & $0.989\,(\pm 0.01)$
        & $\boldsymbol{0.993}\,(\pm 0.01)$ \\

        \bottomrule
    \end{tabular}%
    }
\end{table*}

\begin{figure*}[!t]
\centering
\vspace{-1.0cm}
  \scalebox{1.0}{
  \includegraphics[width=1\textwidth,height=0.6\textwidth]{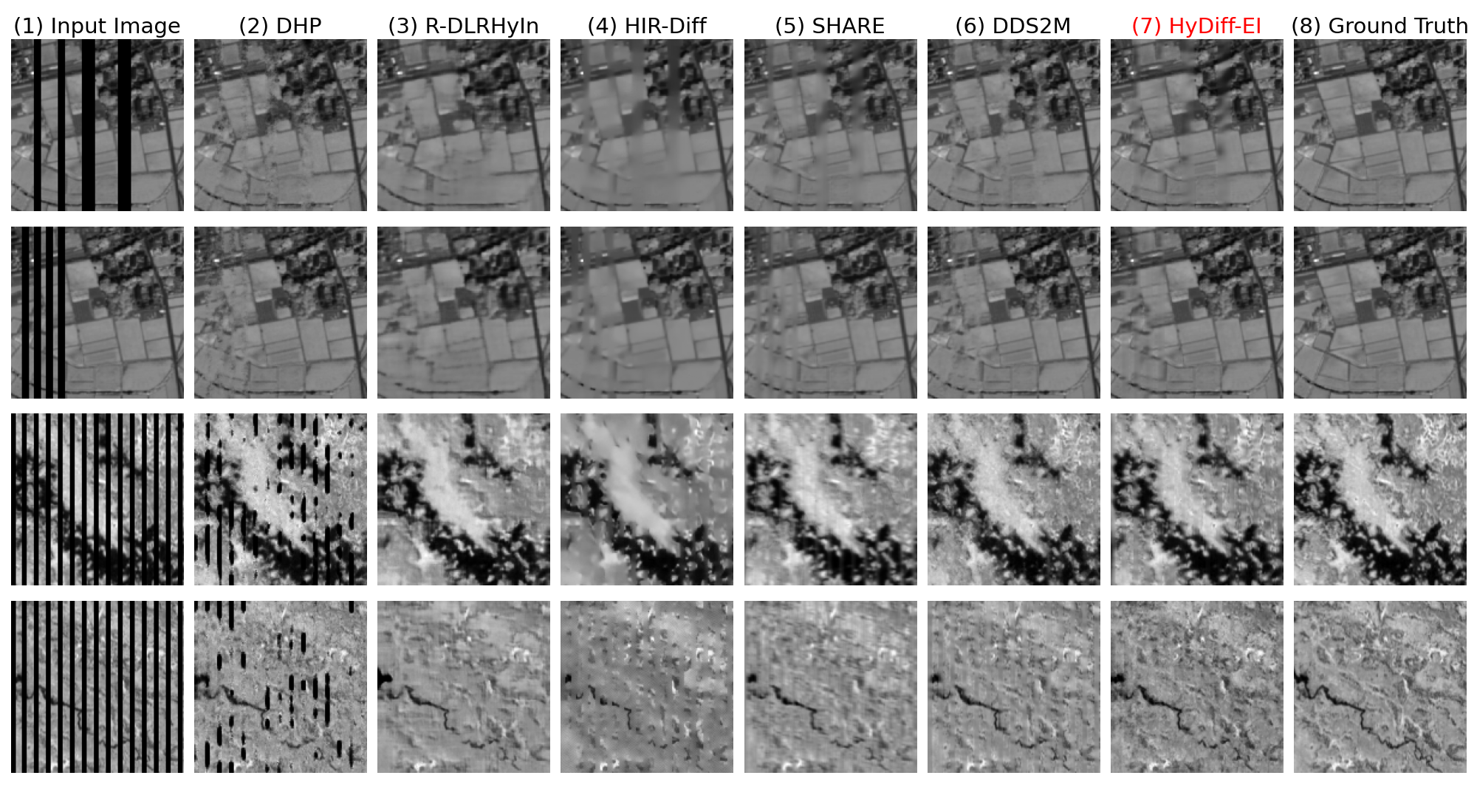}
} 
\vspace{-0.5cm}
  \caption{Visual comparison between the proposed HyDiff-EI algorithm and other HSI inpainting methods on the Chikusei dataset and Botswana dataset (noiseless case). All images are visualized at spectral band 100.}
  \label{HSI_inpainging_noiseless}  
\end{figure*}

\begin{figure*}[t]
  \centering
  % --- First Image (Noisy) ---
  \includegraphics[width=1\textwidth, height=0.6\textwidth]{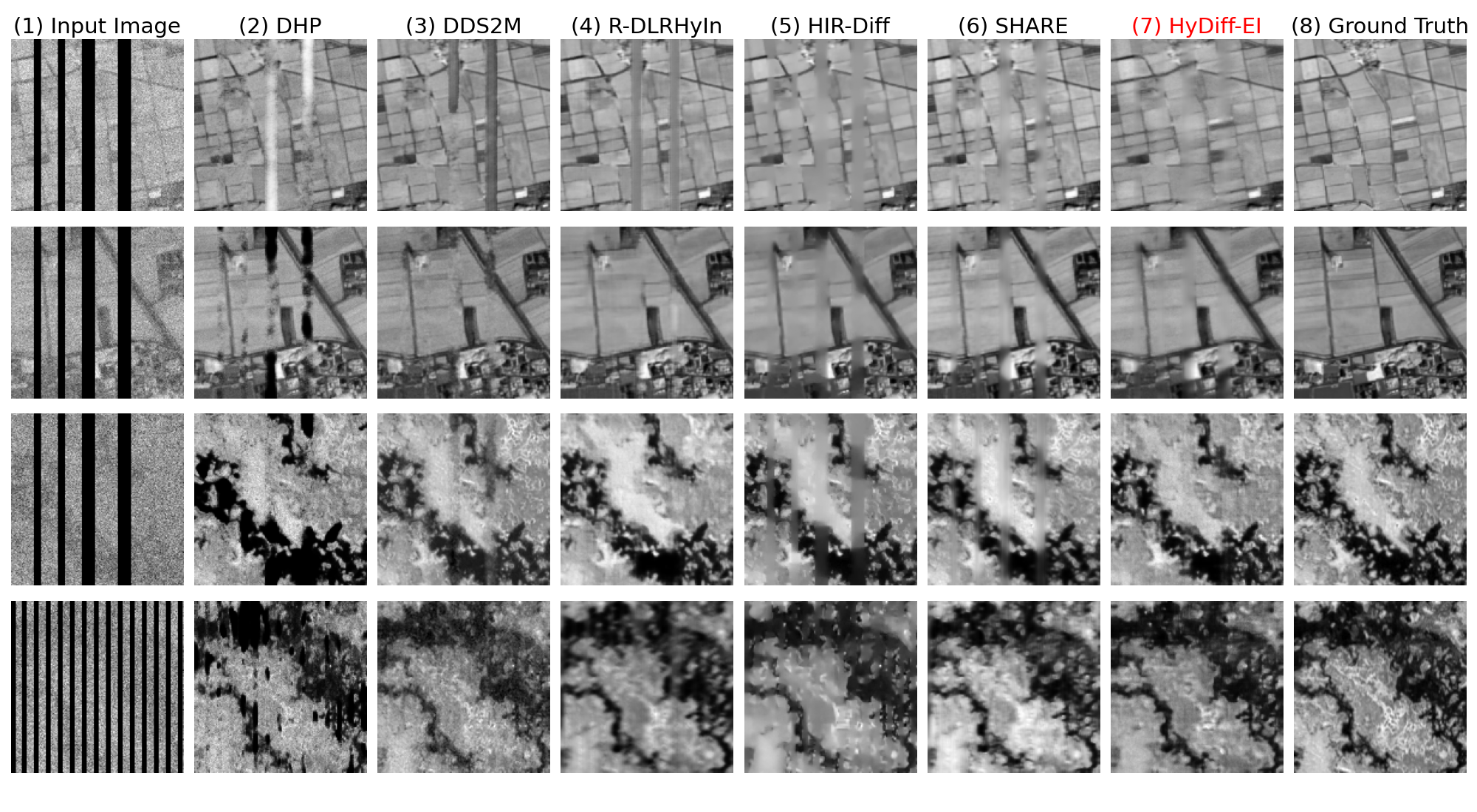}
 \vspace{-0.5cm}
\caption{Visual comparison of the proposed HyDiff-EI algorithm with competing methods on the Chikusei and Botswana datasets under a noise level of $\sigma_y=0.1$. All images are visualized at spectral band 100.}
  \label{HSI_inpainging_noisy}  
  
  % --- Space between figures ---
  \vspace{0.5cm} 
  
  % --- Second Image (EMIT) ---
  \includegraphics[width=1\textwidth, height=0.6\textwidth]{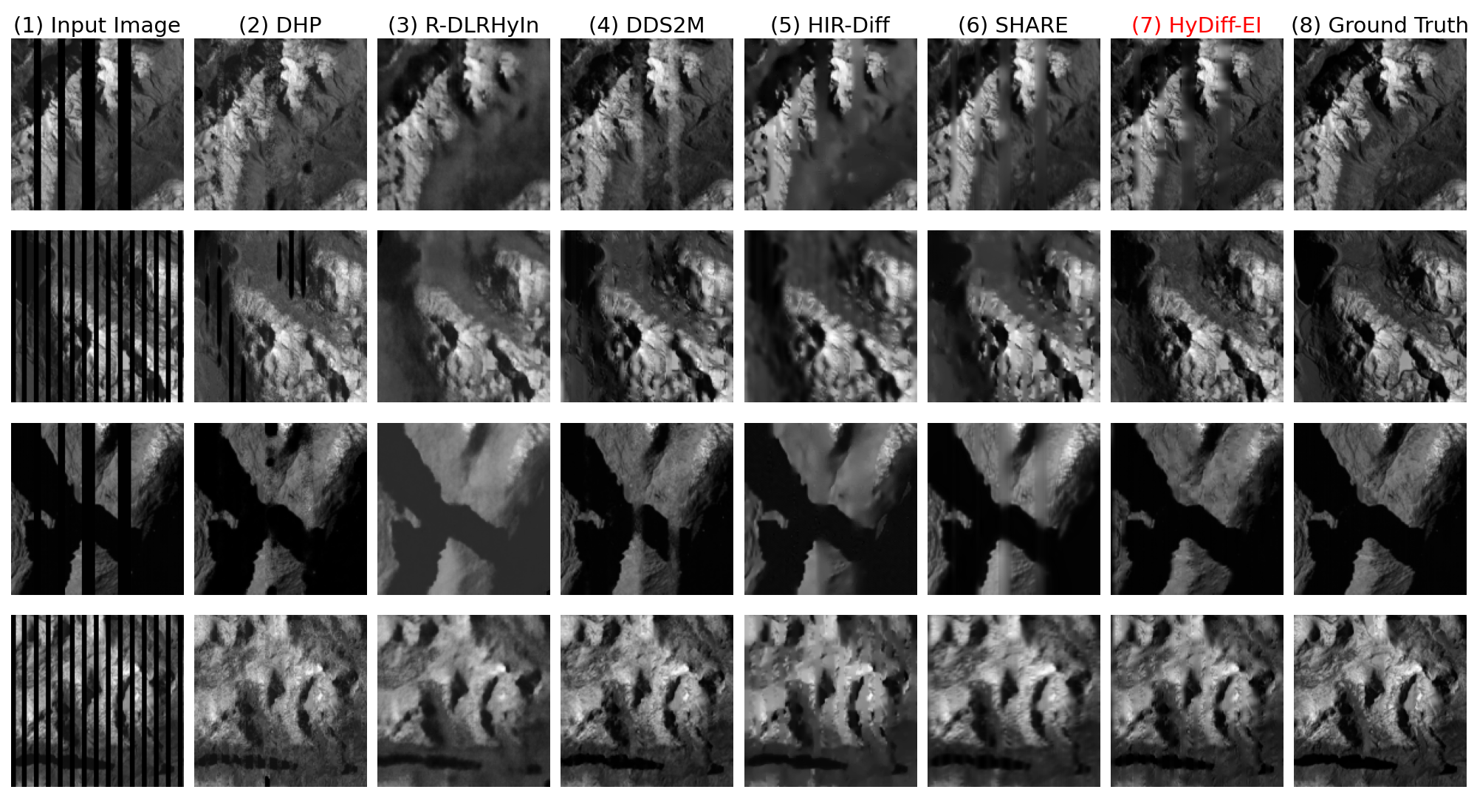}
   \vspace{-0.5cm}
\caption{Visual comparison of the proposed HyDiff-EI algorithm with competing methods on the EMIT dataset. All images are visualized at spectral band 150.}

  \label{HSI_inpainging_EMIT}  

\end{figure*}

We record the mean peak signal-to-noise ratio (MPSNR), mean structural similarity (MSSIM), spectral angle mapper (SAM), and spectral correlation coefficient (SCC) to evaluate HSI inpainting performance. MPSNR and MSSIM mainly assess pixel-level accuracy and spatial structural similarity, while SAM and SCC evaluate spectral fidelity by measuring the angular discrepancy and correlation between reconstructed and ground-truth spectra. The numerical results are reported in Table \ref{Table_HSI_Clean} and \ref{Table_HSI_Noisy}, and the visual results in Figure \ref{HSI_inpainging_noiseless} and \ref{HSI_inpainging_noisy} for noiseless and noisy HSI inpainting, respectively. In both noiseless and noisy cases, HyDiff-EI offers huge improvements over existing methods, effectively filling the missing regions with spectrally consistent and visually smooth content. DDS2M, HIR-Diff, and the proposed HyDiff-EI are three advanced methods that leverage diffusion models. HIR-Diff offers the fastest inference among the three; however, it relies on a pre-trained diffusion model trained on extensive remote-sensing datasets, which limits its generalization ability to unseen data. DDS2M is perhaps the closest method to ours which trains the diffusion model in a self-supervised manner. However, HyDiff-EI is fundamentally different from DDS2M in that: (1) It leverages EI rather than DHP prior. This grants access to a diverse set of virtual measurement operators defined by specific transformations $T$ that may have different null spaces, which is particularly helpful for the inpainting task. (2) While DDS2M relies on DIP, it imposes predefined, hard “black-box" architectural priors on the data with no guarantee of correctness for high-dimensional HSIs. In contrast, Equivariant Imaging (EI) provides a mathematically grounded mechanism to learn missing data by exploiting imaging invariance. (3) Our framework is built upon DDNM \cite{DDNM} which demonstrates superiority than DDRM \cite{DDRM} in the noiseless case.

From Figure \ref{HSI_inpainging_noiseless}, it can be seen that HyDiff-EI better preserves the consistency between the generated content and the surrounding edges than both DDS2M and HIR-Diff. Moreover, it maintains robust inpainting performance even in the extreme case where one-third of the pixels are missing from the input image, as shown in the first test sample in Figure \ref{HSI_inpainging_noiseless}. Note that such missing patterns are rather common in real-world hyperspectral imagers due to the failures of the Scan Line Corrector (SLC) or malfunctions in satellite sensor arrays \footnote{EMIT L1B At-Sensor Calibrated Radiance and Geo-location Data products in \url{https://search.earthdata.nasa.gov/search/} suffer from missing lines/strips.}. Figure \ref{HSI_inpainging_noisy} presents the comparison of different methods in the noisy case with a noise level of $\sigma_y = 0.1$. In this case, DHP struggles or even fails, possibly due to the lack of effective regularization. Surprisingly, R-DLRHyIn achieves higher inpainting quality than DDS2M, although it noticeably smooths sharp edges in the missing regions. This effect is less severe in HIR-Diff, SHARE, and HyDiff-EI, as observed in the first test sample in Figure \ref{HSI_inpainging_noisy}. 
\begin{table*}[!t]
    \centering
    \vspace{-1.0cm}
    \caption{\small
    Comparison between our proposed algorithm and other learning-based
    inpainting algorithms on the Chikusei and Botswana datasets with noise
    levels $\sigma_y=0.1$, $0.2$, and $0.3$. The mean and standard deviation over 20 samples are reported.
    }
    \label{Table_HSI_Noisy}

    \vspace{-0.3cm}
    \setlength{\tabcolsep}{4pt}
    \renewcommand{\arraystretch}{1.15}

    \resizebox{\linewidth}{!}{%
    \begin{tabular}{ccrrrrrrr}
        \toprule
        Noise Level
        & Metric
        & Input
        & DHP \cite{DHP}
        & DDS2M \cite{miao2023dds2m}
        & R-DLRHyIn \cite{niresi2023robust}
        & HIR-Diff \cite{pang2024hir}
        & SHARE \cite{xie2026share}
        & HyDiff-EI (\textcolor{red}{proposed}) \\
        \midrule

        \multirow{4}{*}{$\sigma_y=0.1$}
        & MPSNR $\uparrow$
        & 16.797
        & $27.504\,(\pm 0.77)$
        & $30.683\,(\pm 0.37)$
        & $30.848\,(\pm 0.42)$
        & $31.294\,(\pm 0.39)$
        & $31.468\,(\pm 0.48)$
        & $\boldsymbol{31.784}\,(\pm 0.50)$ \\

        & MSSIM $\uparrow$
        & 0.167
        & $0.734\,(\pm 0.02)$
        & $0.829\,(\pm 0.01)$
        & $0.837\,(\pm 0.01)$
        & $0.867\,(\pm 0.01)$
        & $0.878\,(\pm 0.01)$
        & $\boldsymbol{0.883}\,(\pm 0.01)$ \\

        & SAM ($^\circ$) $\downarrow$
        & --
        & $5.193\,(\pm 2.21)$
        & $17.595\,(\pm 6.98)$
        & $4.719\,(\pm 2.20)$
        & $3.727\,(\pm 0.79)$
        & $3.575\,(\pm 0.74)$
        & $\boldsymbol{3.432}\,(\pm 0.94)$ \\

        & SCC $\uparrow$
        & --
        & $0.963\,(\pm 0.05)$
        & $0.983\,(\pm 0.01)$
        & $0.964\,(\pm 0.05)$
        & $\boldsymbol{0.984}\,(\pm 0.01)$
        & $0.983\,(\pm 0.01)$
        & $0.984\,(\pm 0.02)$ \\

        \midrule

        \multirow{4}{*}{$\sigma_y=0.2$}
        & MPSNR $\uparrow$
        & 14.266
        & $26.898\,(\pm 0.82)$
        & $29.012\,(\pm 0.38)$
        & $28.485\,(\pm 0.42)$
        & $29.570\,(\pm 0.44)$
        & $29.564\,(\pm 0.49)$
        & $\boldsymbol{30.228}\,(\pm 0.62)$ \\

        & MSSIM $\uparrow$
        & 0.154
        & $0.690\,(\pm 0.02)$
        & $0.761\,(\pm 0.01)$
        & $0.755\,(\pm 0.01)$
        & $0.828\,(\pm 0.01)$
        & $0.840\,(\pm 0.01)$
        & $\boldsymbol{0.847}\,(\pm 0.01)$ \\

        & SAM ($^\circ$) $\downarrow$
        & --
        & $6.10\,(\pm 2.45)$
        & $21.30\,(\pm 7.50)$
        & $6.00\,(\pm 2.55)$
        & $4.45\,(\pm 1.05)$
        & $4.27\,(\pm 0.89)$
        & $\boldsymbol{4.05}\,(\pm 0.95)$ \\

        & SCC $\uparrow$
        & --
        & $0.945\,(\pm 0.06)$
        & $0.976\,(\pm 0.02)$
        & $0.945\,(\pm 0.06)$
        & $0.978\,(\pm 0.02)$
        & $0.979\,(\pm 0.01)$
        & $\boldsymbol{0.981}\,(\pm 0.02)$ \\

        \midrule

        \multirow{4}{*}{$\sigma_y=0.3$}
        & MPSNR $\uparrow$
        & 11.998
        & $24.828\,(\pm 1.06)$
        & $27.739\,(\pm 0.40)$
        & $26.332\,(\pm 0.44)$
        & $28.742\,(\pm 0.46)$
        & $\boldsymbol{28.961}\,(\pm 0.51)$
        & $28.610\,(\pm 0.64)$ \\

        & MSSIM $\uparrow$
        & 0.140
        & $0.676\,(\pm 0.03)$
        & $0.731\,(\pm 0.02)$
        & $0.718\,(\pm 0.02)$
        & $0.814\,(\pm 0.03)$
        & $\boldsymbol{0.819}\,(\pm 0.02)$
        & $0.802\,(\pm 0.02)$ \\

        & SAM ($^\circ$) $\downarrow$
        & --
        & $7.35\,(\pm 2.85)$
        & $24.80\,(\pm 8.20)$
        & $7.10\,(\pm 2.90)$
        & $5.20\,(\pm 1.25)$
        & $\boldsymbol{5.08}\,(\pm 0.92)$
        & $5.45\,(\pm 1.15)$ \\

        & SCC $\uparrow$
        & --
        & $0.925\,(\pm 0.07)$
        & $0.962\,(\pm 0.02)$
        & $0.928\,(\pm 0.07)$
        & $0.969\,(\pm 0.03)$
        & $\boldsymbol{0.972}\,(\pm 0.02)$
        & $0.966\,(\pm 0.03)$ \\

        \bottomrule
    \end{tabular}%
    }
\end{table*}

\clearpage
\newpage

\clearpage

\newpage

\begin{table*}[!t]
    \centering
    \caption{\small
    Comparison between the proposed HyDiff-EI algorithm and other
    learning-based inpainting methods on the EMIT dataset. The mean and
    standard deviation over 20 samples are reported.
    }
    \label{Table_EMIT}

    \vspace{-0.3cm}
    \setlength{\tabcolsep}{4pt}
    \renewcommand{\arraystretch}{1.15}

    \resizebox{\linewidth}{!}{%
    \begin{tabular}{crrrrrrr}
        \toprule
        Methods
        & Input
        & DHP \cite{DHP}
        & R-DLRHyIn \cite{niresi2023robust}
        & DDS2M \cite{miao2023dds2m}
        & HIR-Diff \cite{pang2024hir}
        & SHARE \cite{xie2026share}
        & HyDiff-EI (\textcolor{red}{proposed}) \\
        \midrule

        MPSNR $\uparrow$
        & 22.517
        & $31.319\,(\pm 0.74)$
        & $33.227\,(\pm 0.61)$
        & $33.358\,(\pm 0.54)$
        & $33.592\,(\pm 0.48)$
        & $34.673\,(\pm 0.50)$
        & $\boldsymbol{35.482}\,(\pm 0.49)$ \\

        MSSIM $\uparrow$
        & 0.198
        & $0.880\,(\pm 0.05)$
        & $0.878\,(\pm 0.04)$
        & $0.905\,(\pm 0.01)$
        & $0.912\,(\pm 0.01)$
        & $0.920\,(\pm 0.01)$
        & $\boldsymbol{0.933}\,(\pm 0.01)$ \\

        SAM ($^\circ$) $\downarrow$
        & --
        & $3.601\,(\pm 0.43)$
        & $4.397\,(\pm 0.45)$
        & $3.147\,(\pm 0.47)$
        & $4.454\,(\pm 0.52)$
        & $2.782\,(\pm 0.39)$
        & $\boldsymbol{2.095}\,(\pm 0.42)$ \\

        SCC $\uparrow$
        & --
        & $0.968\,(\pm 0.00)$
        & $0.963\,(\pm 0.01)$
        & $0.974\,(\pm 0.01)$
        & $0.965\,(\pm 0.00)$
        & $0.978\,(\pm 0.01)$
        & $\boldsymbol{0.982}\,(\pm 0.01)$ \\

        \bottomrule
    \end{tabular}%
    }
\end{table*}

\begin{figure*}[!t]
    \centering
    \begin{minipage}{0.49\textwidth}
        \centering
        \includegraphics[width=0.98\linewidth]{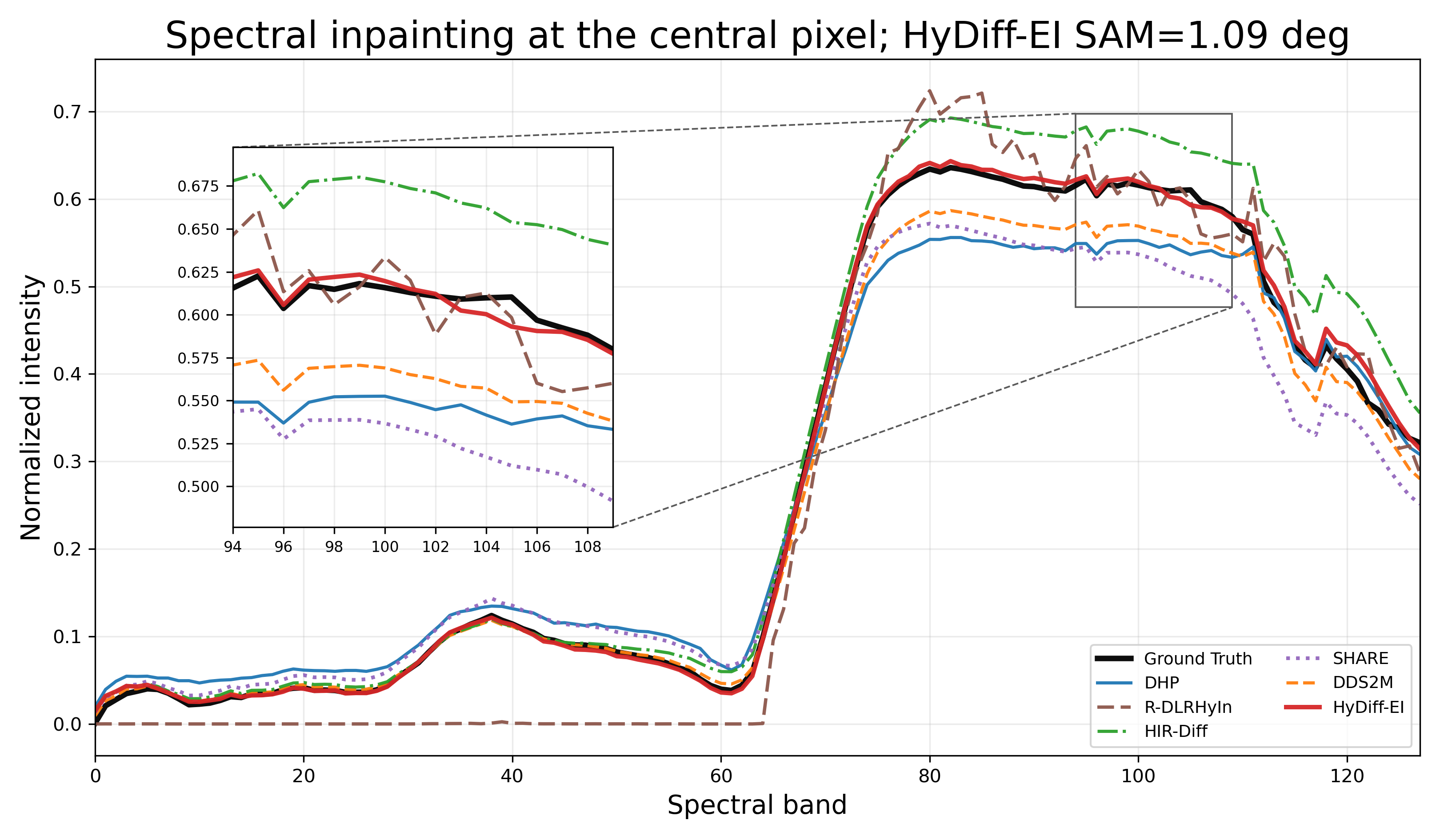}
        \vspace{-0.15cm}
        \centerline{\small (a) Noiseless case: central pixel}
    \end{minipage}\hfill
    \begin{minipage}{0.49\textwidth}
        \centering
        \includegraphics[width=0.98\linewidth]{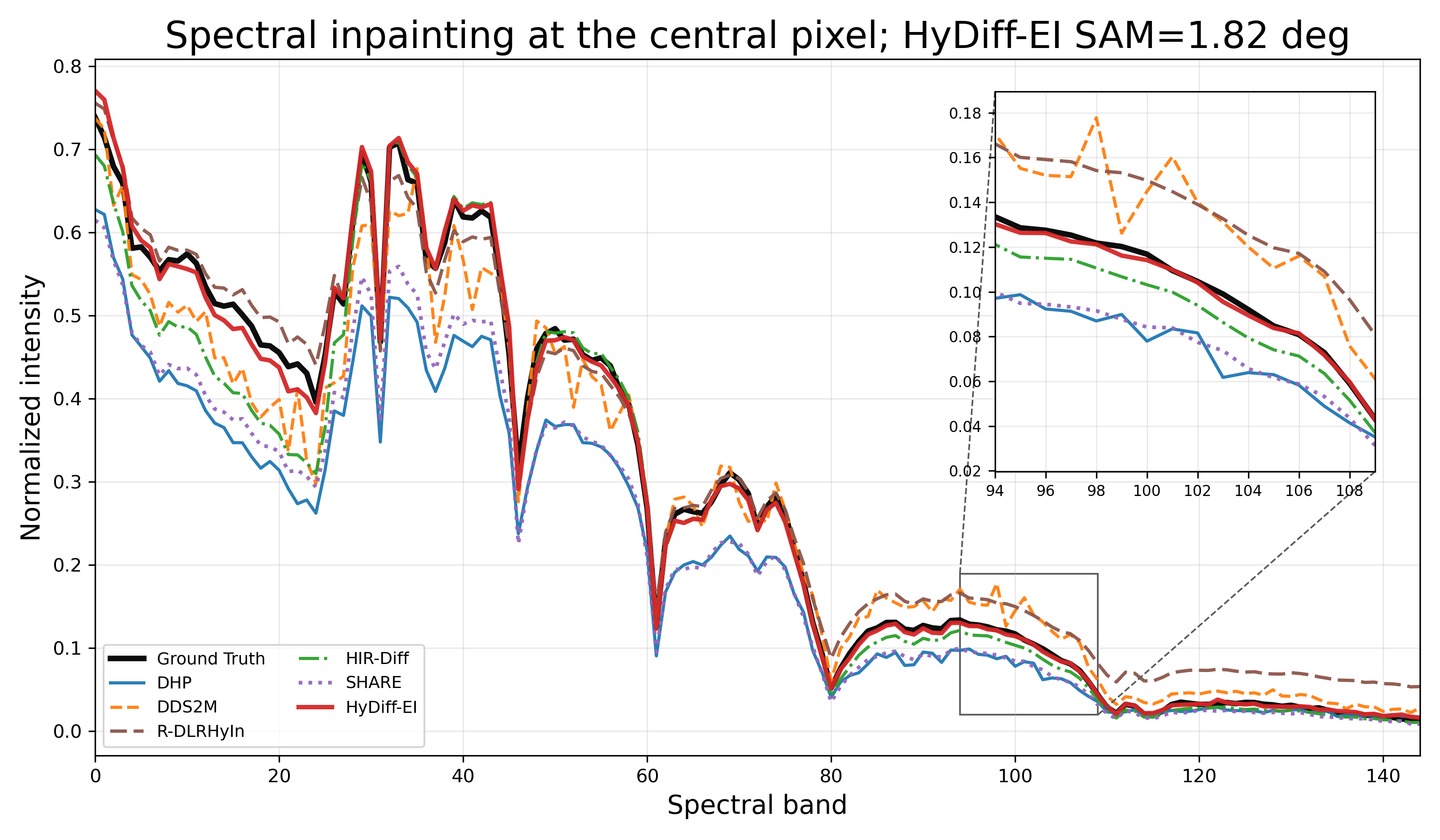}
        \vspace{-0.15cm}
        \centerline{\small (b) Noisy case: central pixel}
    \end{minipage}

    \vspace{0.5cm}

    \begin{minipage}{0.49\textwidth}
        \centering
        \includegraphics[width=0.98\linewidth]{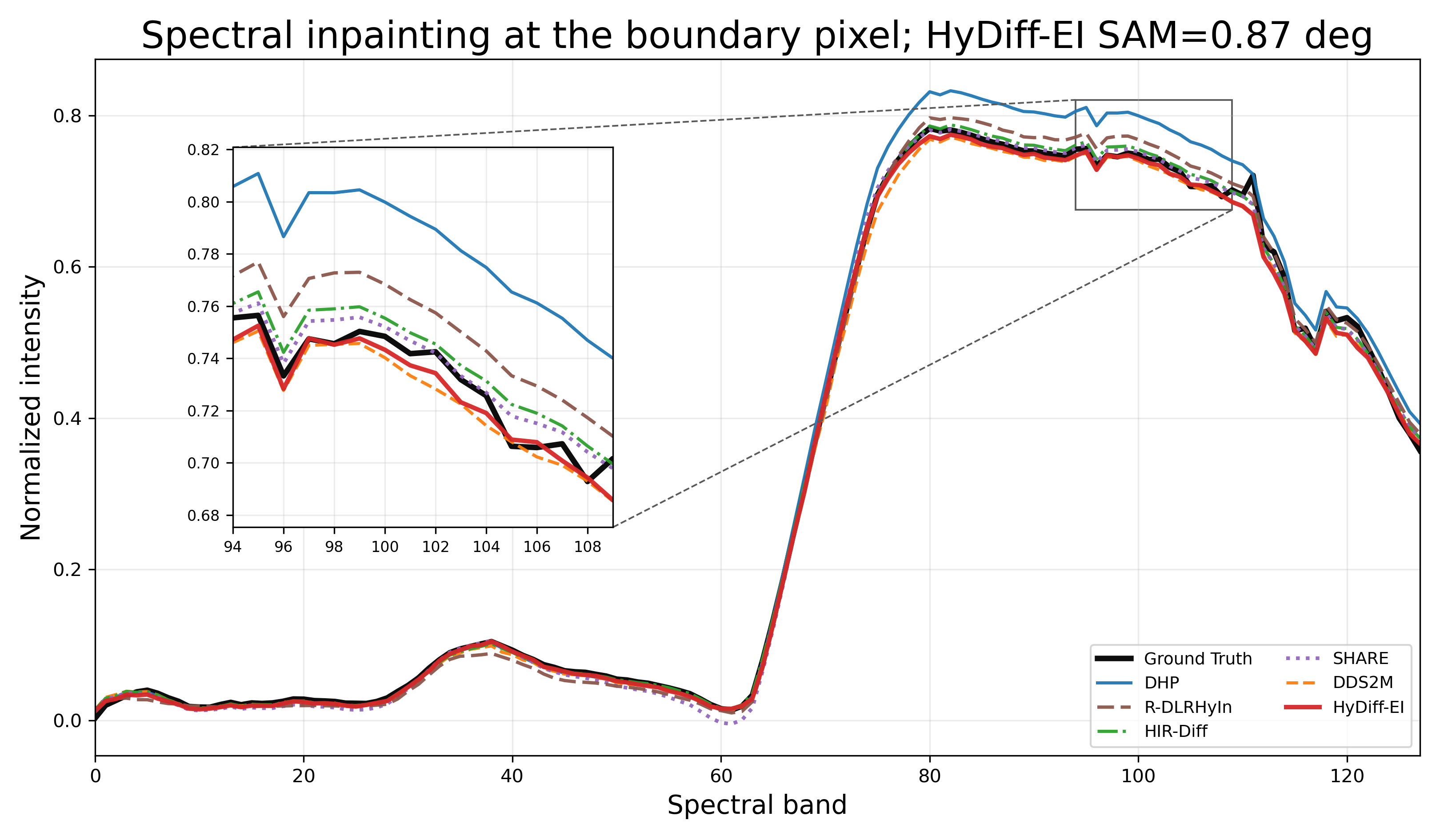}
        \vspace{-0.15cm}
        \centerline{\small (c) Noiseless case: boundary pixel}
    \end{minipage}\hfill
    \begin{minipage}{0.49\textwidth}
        \centering
        \includegraphics[width=0.98\linewidth]{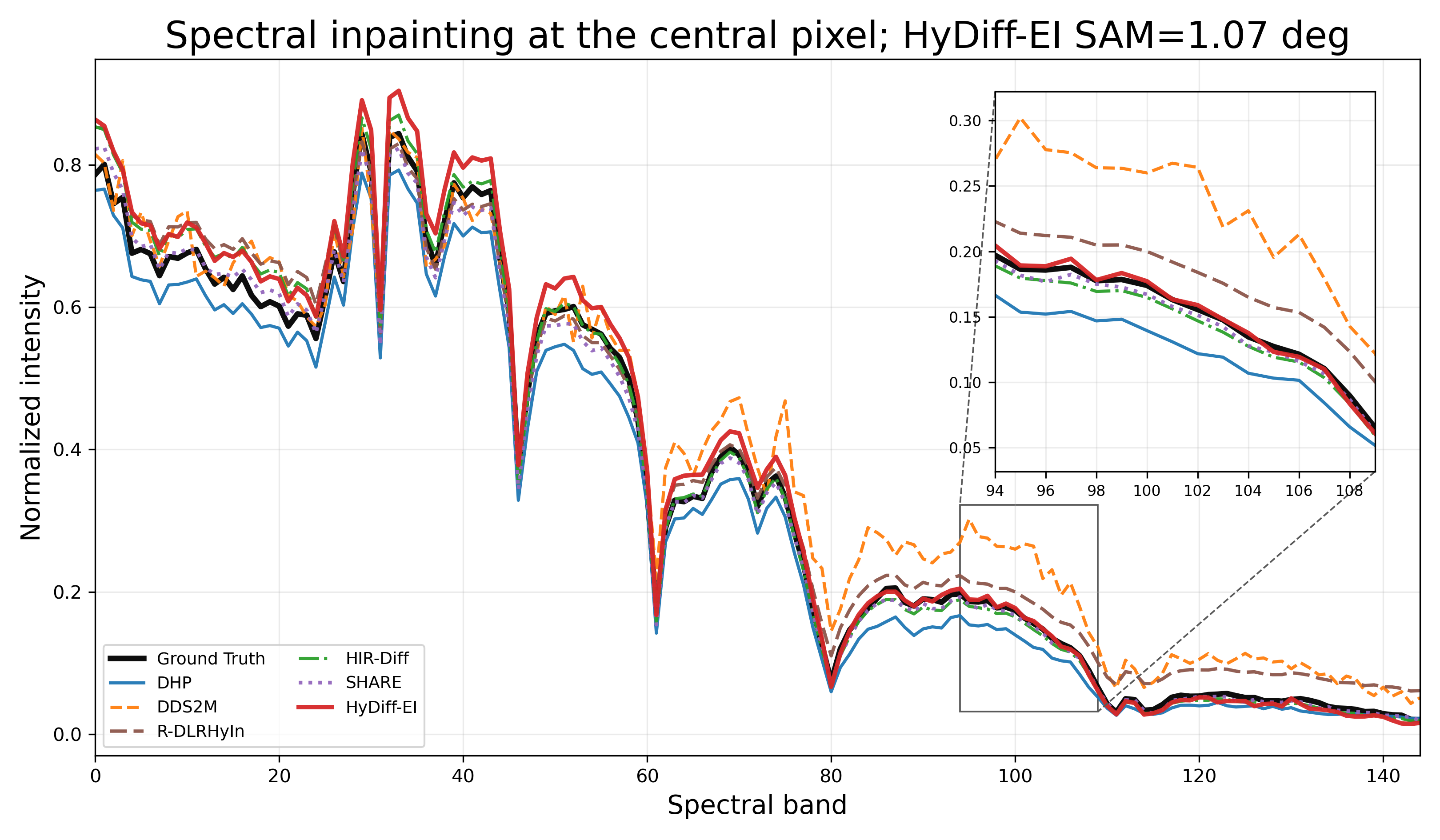}
        \vspace{-0.15cm}
        \centerline{\small (d) Noisy case: boundary pixel}
    \end{minipage}

    \vspace{0.2cm}

\caption{Comparison of reconstructed spectral signatures at representative inpainted pixels in Chikusei test samples under noiseless and noisy conditions. The left column shows the noiseless results, while the right column shows the noisy results.}
\label{fig:spectral_curves_comparison}

\end{figure*}
From Table \ref{Table_HSI_Noisy}, it can be observed that as the noise level increases, the performance gap between HyDiff-EI and SHARE gradually narrows. HyDiff-EI achieves the best overall performance at $\sigma_y=0.1$ and $0.2$, whereas SHARE surpasses HyDiff-EI at $\sigma_y=0.3$ in terms of MPSNR, MSSIM, SAM, and SCC. This suggests that SHARE exhibits stronger robustness under severe noise, while HyDiff-EI is more effective under low-to-moderate noise levels. These results highlight the importance of incorporating effective regularization priors into self-supervised reconstruction networks. Nevertheless, the performance degradation of HyDiff-EI under severe noise also indicates the need for further noise-aware regularization. \\
\noindent \textbf{Spectral curve comparison.} In addition to the image-level evaluation,  Fig.~\ref{fig:spectral_curves_comparison} further compares the reconstructed spectral signatures at representative missing pixels under both noiseless and noisy settings, including central and boundary locations inside the missing regions. In the noiseless case, HyDiff-EI follows the ground-truth spectra more closely than the competing methods at both the central and boundary pixels, especially in the zoomed-in wavelength intervals. Under the noisy setting, HyDiff-EI still preserves the spectral shape more faithfully, while several methods show larger angular deviations or local spectral distortions. These observations are consistent with the strong SAM and SCC performance reported in Table~\ref{Table_HSI_Noisy}, indicating that HyDiff-EI better preserves material-dependent spectral signatures instead of only improving pixel-level visual quality. This capability is important for downstream applications such as hyperspectral image classification. \\
\noindent \textbf{Validate on Real-time EMIT Data.} Figure~\ref{HSI_inpainging_EMIT} and Table~\ref{Table_EMIT} present the visual and quantitative inpainting results for an image sample from the EMIT mission ~\cite{EMIT_Mission}.\footnote{Image product acquisition date: September 4, 2025.} In the visual comparison, HyDiff-EI preserves more fine-scale spatial structures in the mountain regions within the missing strips than DDS2M, HIR-Diff and SHARE. This advantage is quantitatively confirmed in Table~\ref{Table_EMIT}, where HyDiff-EI achieves the best performance on all four metrics, with an MPSNR of $35.482$, an MSSIM of $0.933$, a SAM of $2.095^\circ$, and an SCC of $0.982$. Unlike the DHP families (DHP, DDS2M and R-DLRHyIn) which leverage the network structure as priors, HyDiff-EI shows that incorporating an equivariant prior can significantly improve inpainting quality. Moreover, both HyDiff-EI and DDS2M produce less over-smoothed spectral content than HIR-Diff for this image sample, further highlighting the advantage of self-supervised diffusion approaches over fully supervised methods. 
\section{Ablation Studies} \label{section_ablations}
\noindent \textbf{Does Diffusion help?} To verify whether the introduced diffusion framework contributes to this success, Table~\ref{Ablation_EI} presents the inpainting performance of EI (equivalently, Hyper-EI in \cite{Hyper-EI}) with and without diffusion in the noiseless setting. For the noisy case, the results are compared with Robust-EI \cite{REI}, an extension of EI designed for noisy measurements. The reported values represent averages over 20 independent runs with different random seeds. From Table \ref{Ablation_EI}, it is observed that HyDiff-EI enjoys a steady performance gain in the noiseless case. As the noise level increases, the MPSNR gap between Robust-EI and HyDiff-EI decreases, whereas the MSSIM improves, indicating that HyDiff-EI is able to preserve high-frequency details and enhances perceptual quality. The advantage of diffusion-based reconstruction lies in its ability to sample from the posterior distribution rather than producing a single point estimate such as the MMSE (minimum mean squared error) solution. Unlike MMSE-like estimators, which tend to yield over-smoothed results by averaging plausible solutions, diffusion-based EI is able to inpaint with sharp and more realistic structural details.
\begin{table}[h]
    \centering
    \caption{Performance of EI with and without diffusion on the HSI inpainting task under varying noise levels. Since EI is not originally designed for noisy measurements, Robust-EI is included as a reference for the noisy inpainting setting.}
    \label{Ablation_EI}
    \resizebox{\linewidth}{!}{
        \begin{tabular}{clccc}
        \toprule
        Noise Level & Metric & EI & Robust-EI & HyDiff-EI \\
        \midrule
        $\sigma_y = 0$ & MPSNR $\uparrow$ & $34.798 \, (\pm 0.34)$ & $34.806 \, (\pm 0.35)$ & $\boldsymbol{36.527} \, (\pm 0.36)$ \\
        & MSSIM $\uparrow$ & $0.895 \, (\pm 0.01)$ & $0.895 \, (\pm 0.01)$ & $\boldsymbol{0.904} \, (\pm 0.01)$ \\
        \midrule
        $\sigma_y = 0.1$ & MPSNR $\uparrow$ & $32.381 \, (\pm 0.59)$ & $\boldsymbol{34.628} \, (\pm 0.74)$ & ${34.242} \, (\pm 0.68)$ \\
        & MSSIM $\uparrow$ & $0.880 \, (\pm 0.01)$ & $0.892 \, (\pm 0.01)$ & $\boldsymbol{0.893} \, (\pm 0.01)$ \\
        \midrule
        $\sigma_y = 0.2$ & MPSNR $\uparrow$ & $31.660 \, (\pm 0.77)$ & $33.121 \, (\pm 0.70)$ & $\boldsymbol{33.179} \, (\pm 0.67)$ \\
        & MSSIM $\uparrow$ & $0.867 \, (\pm 0.01)$ & $0.874 \, (\pm 0.01)$ & $\boldsymbol{0.878} \, (\pm 0.01)$ \\
        \midrule
        $\sigma_y = 0.3$ & MPSNR $\uparrow$ & $29.811 \, (\pm 1.23)$ & $31.518 \, (\pm 0.81)$ & $\boldsymbol{31.655} \, (\pm 0.85)$ \\
        & MSSIM $\uparrow$ & $0.838 \, (\pm 0.02)$ & $0.847 \, (\pm 0.01)$ & $\boldsymbol{0.861} \, (\pm 0.01)$ \\
        \bottomrule
        \end{tabular}
    }
\end{table}

\noindent \textbf{Time Efficiency.} 
Table \ref{Time_Comparison} provides a time-efficiency evaluation of HyDiff-EI against other competing methods, with all tests conducted on a single NVIDIA Tesla V100 GPU using a $144 \times 144 \times 128$ Chikusei image sample. Runtime is reported in $\mathrm{s/image}$. The results reveal a clear quality and efficiency trade-off across the evaluated methods. DHP and its extension R-DLRHyIn are the most lightweight solutions: DHP achieves the shortest runtime of $28.12~\mathrm{s/image}$, while R-DLRHyIn substantially improves the reconstruction quality with only a modest increase in runtime to $31.79~\mathrm{s/image}$. HIR-Diff requires $53.33~\mathrm{s/image}$ despite using the pre-trained DDPM-CD model \cite{bandara2022ddpm_CD}, which can be attributed to its relatively large model complexity and parameter count. SHARE has a comparable runtime of approximately $52~\mathrm{s/image}$; its additional cost may be partly attributed to the larger DASA block and the extra network forward pass required for evaluating the SURE loss. DDS2M is substantially more expensive, requiring $178.43~\mathrm{s/image}$, since its diffusion-based optimization repeatedly updates the DIP parameters over numerous diffusion backward steps. Consequently, DDS2M requires approximately $4.6$ times the runtime of HyDiff-EI. In contrast, the proposed HyDiff-EI achieves the best restoration performance while requiring only $38.54~\mathrm{s/image}$. This corresponds to a modest runtime overhead of approximately $37\%$ relative to the DHP baseline, while remaining approximately $26\%$ faster than SHARE and more than four times faster than DDS2M. The additional computational cost of HyDiff-EI arises from the equivariant imaging regularization and diffusion-based backward updates. Overall, these results demonstrate that HyDiff-EI provides a favorable balance between reconstruction quality and computational efficiency.

\begin{table}[h]
    \centering
    \caption{Time efficiency and performance comparison evaluated on a
    $144 \times 144 \times 128$ Chikusei sample.}
    \label{Time_Comparison}
    \resizebox{\linewidth}{!}{
        \begin{tabular}{lcccccc}
        \toprule
        \textbf{Method} & DHP & R-DLRHyIn & HIR-Diff (pre-trained)
        & DDS2M & SHARE & \textbf{HyDiff-EI} \\
        \midrule
        \textbf{Time (s/image)}
        & \textbf{28.12}
        & 31.79
        & 53.33
        & 178.43
        & 52.00
        & \textbf{38.54} \\

        \textbf{MPSNR (dB)}$\uparrow$
        & 28.425
        & 32.464
        & 32.812
        & 33.258
        & 34.108
        & $\boldsymbol{34.326}$ \\

        \textbf{MSSIM}$\uparrow$
        & 0.842
        & 0.889
        & 0.893
        & 0.905
        & 0.911
        & $\boldsymbol{0.917}$ \\
        \bottomrule
        \end{tabular}
    }
    
    \vspace{2mm}
    
    \raggedright
    \footnotesize
\end{table}

\noindent \textbf{Effect of $\alpha_{EI}$ on the Inpainting Performance.} 
Table \ref{EI_weight_ablation} evaluates the sensitivity of HyDiff-EI to the equivariance strength hyperparameter, $\alpha_{EI}$, across different noise levels ($\sigma_y$). This parameter controls the relative weight of the EI-consistency loss relative to the data-fidelity loss. As observed, the optimal balance is consistently reached at $\alpha_{EI} = 1$. Removing the EI constraint entirely ($\alpha_{EI} = 0$) causes a severe performance collapse, particularly under noisy conditions. Conversely, setting $\alpha_{EI}$ too high (e.g., $\alpha_{EI} = 2$) results in reduced inpainting performance. Hence, we set $\alpha_{EI} = 1$ as the default for all experiments to achieve the optimal trade-off.

\begin{table}[h]
    \centering
    \caption{Quantitative evaluations of HyDiff-EI with different $\alpha_{EI}$.}
    \label{EI_weight_ablation}
    \resizebox{\linewidth}{!}{
        \begin{tabular}{llccccc}
        \toprule
        Noise Level & Metric & $\alpha_{EI}=0$ & $0.1$ & $0.5$ & $1$ & $2$ \\
        \midrule
        $\sigma_y = 0$
        & MPSNR $\uparrow$ & $28.436$ & $30.127$ & $33.495$ & $\boldsymbol{35.762}$ & $33.355$ \\
        & MSSIM $\uparrow$ & $0.753$ & $0.879$ & $0.906$ & $\boldsymbol{0.932}$ & $0.904$ \\
        \midrule
        $\sigma_y = 0.2$
        & MPSNR $\uparrow$ & $15.618$ & $27.356$ & $30.642$ & $\boldsymbol{32.071}$ & $28.259$ \\
        & MSSIM $\uparrow$ & $0.161$ & $0.714$ & $0.839$ & $\boldsymbol{0.891}$ & $0.732$ \\
        \bottomrule
        \end{tabular}
    }
\end{table}
\noindent \textbf{Effect of Different Types of Transformations.}
The selection of an appropriate transformation group is crucial for the HS inpainting task. Table \ref{tab:transformation_ablation} evaluates six transformations from the Deep Inverse library \cite{tachella2025deepinverse} within our HyDiff-EI framework. Shifting and Rotation achieve the highest and most stable inpainting performance, suggesting that spatial equivariance that preserves local structures and spectral signatures are particularly effective for HSI inpainting. This is consistent with the nature of the inpainting problem, where missing regions should be reconstructed using surrounding spatial context while maintaining spectral consistency across bands. We leave the investigation of bespoke spectral-domain transformations, tailored specifically to the physical properties of HSIs, as an important direction for future research.
\begin{table}[h]
    \centering
    \caption{Quantitative evaluations of HyDiff-EI with different types of transformations. The first four are basic transformations, and the last two are projection-based.}
    \label{tab:transformation_ablation}
    \begin{tabular}{lcc}
    \toprule
    Transformation & MPSNR $\uparrow$ & MSSIM $\uparrow$ \\
    \midrule
    Shifting  & $\boldsymbol{34.602} (\pm 0.39)$ & $\boldsymbol{0.926} (\pm 0.01)$ \\
    Rotation   & $34.439 (\pm 0.32)$ & $0.922 (\pm 0.01)$ \\
    Scaling    & $33.477 (\pm 0.25)$ & $0.913 (\pm 0.01)$ \\
    Reflection & $28.334 (\pm 0.72)$ & $0.836 (\pm 0.02)$ \\
    Similarity & $33.161 (\pm 0.20)$ & $0.901 (\pm 0.01)$ \\
    Affine     & $32.104 (\pm 0.47)$ & $0.898 (\pm 0.01)$ \\
    \bottomrule
    \end{tabular}
\end{table}

\section{Conclusion} \label{conclusion}
\noindent We present a self-supervised diffusion probabilistic model solution for the HSI inpainting problem called the HyDiff-EI algorithm. In detail, we develop an untrained EI network under the diffusion framework which can produce realistic and high-quality HSI reconstructions. The proposed HyDiff-EI algorithm exploits the strong regularization capability of the equivariant prior and leverages the high-level hierarchical information of diffusion models. Extensive experiments on the Chikusei, Botswana, and EMIT datasets demonstrate that HyDiff-EI achieves superior overall performance in noiseless and low-to-moderate noise settings, while remaining competitive under severe noise. The runtime comparison further demonstrates that HyDiff-EI achieves a favorable balance between restoration quality and computational efficiency, highlighting the practical advantages of combining equivariant priors with self-supervised diffusion reconstruction. The investigation of more effective noise-aware regularization and physically meaningful transformation groups along the spectral dimension of hyperspectral images remains an important direction for future work.

\bibliographystyle{IEEEtran}
\bibliography{main}

\appendices

\allowdisplaybreaks
\appendix
\section*{\centering {Supplementary Material: Loss Function Details}} 
\label{Appendix_loss} 

We provide a detailed derivation of the loss function \eqref{EI_loss} presented in the main text. The objective of the EI network, $f_{\theta}^{(t)}(\boldsymbol{y})$, is to estimate the clean data $\boldsymbol{x}_0$ from the observation $\boldsymbol{y}$:
\begin{equation}
\begin{aligned} \label{DIP_objective}
 \max_{\theta} \mathbb{E}\left[\log p_{\theta}(\boldsymbol{x}_0|\boldsymbol{y})\right], \quad\text{i.e.,}\quad L = \min_{\theta} \mathbb{E}\left[-\log p_{\theta}(\boldsymbol{x}_0|\boldsymbol{y})\right].
\end{aligned}
\end{equation}

Instead of directly solving \eqref{DIP_objective}, we borrow concepts from Denoising Diffusion Probabilistic Models (DDPMs) to construct a variational upper bound:
\begin{align}
  -\log p_{\theta}(\boldsymbol{x}_0|\boldsymbol{y})
  &\le -\log p_{\theta}(\boldsymbol{x}_0|\boldsymbol{y})
  + \mathbb{D}_{KL}\!\left(q(\boldsymbol{x}_{1:T}|\boldsymbol{x}_0)\right. \notag\\
&\qquad\qquad\left.
\| p_{\theta}(\boldsymbol{x}_{1:T}|\boldsymbol{x}_0,\boldsymbol{y})\right) \notag\\
&= -\log p_{\theta}(\boldsymbol{x}_0|\boldsymbol{y})
 + \int_{-\infty}^{+\infty} q(\boldsymbol{x}_{1:T}|\boldsymbol{x}_0) \notag\\
&\qquad \times
\log\frac{q(\boldsymbol{x}_{1:T}|\boldsymbol{x}_0)}
{p_{\theta}(\boldsymbol{x}_{1:T}|\boldsymbol{x}_0,\boldsymbol{y})}
d\boldsymbol{x}_{1:T} \notag\\
&= -\log p_{\theta}(\boldsymbol{x}_0|\boldsymbol{y})
 + \mathbb{E}\!\left[
\log\frac{q(\boldsymbol{x}_{1:T}|\boldsymbol{x}_0)}
{p_{\theta}(\boldsymbol{x}_{0:T}|\boldsymbol{y})}
\right. \notag\\
&\qquad\qquad\left.
+ \log p_{\theta}(\boldsymbol{x}_0|\boldsymbol{y})\right] \notag\\
&= \mathbb{E}\!\left[
\log\frac{q(\boldsymbol{x}_{1:T}|\boldsymbol{x}_0)}
{p_{\theta}(\boldsymbol{x}_{0:T}|\boldsymbol{y})}
\right] \notag\\
&\triangleq L_{ub},
\end{align}

\noindent where $\mathbb{D}_{KL}$ denotes the Kullback-Leibler (KL) divergence. Bayes' theorem is applied in the third line, allowing the $\log p_{\theta} (\boldsymbol{x}_0|\boldsymbol{y})$ term to cancel out. Denoting this resulting upper bound as $L_{ub}$, we expand it as follows:
\begin{align}
L_{ub}
&= \mathbb{E} \Bigg[
\log\frac{q(\boldsymbol{x}_{1:T}|\boldsymbol{x}_0)}
{p_{\theta}(\boldsymbol{x}_{0:T}|\boldsymbol{y})}
\Bigg]\\
&= \mathbb{E} \Bigg[
\log\frac{
q(\boldsymbol{x}_T|\boldsymbol{x}_{T-1})
q(\boldsymbol{x}_{T-1}|\boldsymbol{x}_{T-2}) \cdots
q(\boldsymbol{x}_1|\boldsymbol{x}_0)}
{p_{\theta}(\boldsymbol{x}_0|\boldsymbol{x}_1) \cdots
p_{\theta}(\boldsymbol{x}_{T-1}|\boldsymbol{x}_T)
p_{\theta}(\boldsymbol{x}_T|\boldsymbol{y})}
\Bigg] \\
&= \mathbb{E} \Bigg[
\log\frac{1}{p_{\theta}(\boldsymbol{x}_T|\boldsymbol{y})}
+ \sum_{t=2}^{T}
\log \frac{q(\boldsymbol{x}_t|\boldsymbol{x}_{t-1})}
{p_{\theta}(\boldsymbol{x}_{t-1}|\boldsymbol{x}_t)}
\nonumber\\
&\qquad\qquad
+ \log \frac{q(\boldsymbol{x}_1|\boldsymbol{x}_0)}
{p_{\theta}(\boldsymbol{x}_0|\boldsymbol{x}_1)}
\Bigg] \\
&= \mathbb{E} \Bigg[
\log\frac{1}{p_{\theta}(\boldsymbol{x}_T|\boldsymbol{y})}
+ \sum_{t=2}^{T}
\log
\frac{
\frac{
q(\boldsymbol{x}_{t-1}|\boldsymbol{x}_{t},\boldsymbol{x}_0)
q(\boldsymbol{x}_t|\boldsymbol{x}_0)}
{q(\boldsymbol{x}_{t-1}|\boldsymbol{x}_0)}
}
{p_{\theta}(\boldsymbol{x}_{t-1}|\boldsymbol{x}_t)}
\nonumber\\
&\qquad\qquad
+ \log \frac{q(\boldsymbol{x}_1|\boldsymbol{x}_0)}
{p_{\theta}(\boldsymbol{x}_0|\boldsymbol{x}_1)}
\Bigg] \\
&= \mathbb{E} \Bigg[
\log\frac{1}{p_{\theta}(\boldsymbol{x}_T|\boldsymbol{y})}
+ \sum_{t=2}^{T}
\log \frac{q(\boldsymbol{x}_{t-1}|\boldsymbol{x}_t,\boldsymbol{x}_0)}
{p_{\theta}(\boldsymbol{x}_{t-1}|\boldsymbol{x}_t)}
\nonumber\\
&\qquad\qquad
+ \sum_{t=2}^{T}
\log \frac{q(\boldsymbol{x}_t|\boldsymbol{x}_0)}
{q(\boldsymbol{x}_{t-1}|\boldsymbol{x}_0)}
+ \log \frac{q(\boldsymbol{x}_1|\boldsymbol{x}_0)}
{p_{\theta}(\boldsymbol{x}_0|\boldsymbol{x}_1)}
\Bigg] \\
&= \mathbb{E} \Bigg[
\log\frac{1}{p_{\theta}(\boldsymbol{x}_T|\boldsymbol{y})}
+ \sum_{t=2}^{T}
\log \frac{q(\boldsymbol{x}_{t-1}|\boldsymbol{x}_t,\boldsymbol{x}_0)}
{p_{\theta}(\boldsymbol{x}_{t-1}|\boldsymbol{x}_t)}
\nonumber\\
&\qquad\qquad
+ \log \frac{q(\boldsymbol{x}_T|\boldsymbol{x}_0)}
{q(\boldsymbol{x}_1|\boldsymbol{x}_0)}
+ \log \frac{q(\boldsymbol{x}_1|\boldsymbol{x}_0)}
{p_{\theta}(\boldsymbol{x}_0|\boldsymbol{x}_1)}
\Bigg] \\
&= \mathbb{E} \Bigg[
\log\frac{q(\boldsymbol{x}_T|\boldsymbol{x}_0)}
{p_{\theta}(\boldsymbol{x}_T|\boldsymbol{y})}
+ \sum_{t=2}^{T}
\log \frac{q(\boldsymbol{x}_{t-1}|\boldsymbol{x}_t,\boldsymbol{x}_0)}
{p_{\theta}(\boldsymbol{x}_{t-1}|\boldsymbol{x}_t)}
\nonumber\\
&\qquad\qquad
- \log p_{\theta}(\boldsymbol{x}_0|\boldsymbol{x}_1)
\Bigg]\\
&= \mathbb{E} \Bigg[
\underbrace{
\mathbb{D}_{KL}(q(\boldsymbol{x}_T|\boldsymbol{x}_0)
\| p_{\theta}(\boldsymbol{x}_T|\boldsymbol{y}))}_{L_T}
\nonumber\\
&\qquad\qquad
+ \sum_{t=2}^{T}
\underbrace{
\mathbb{D}_{KL}(q(\boldsymbol{x}_{t-1}|\boldsymbol{x}_t,\boldsymbol{x}_0)
\| p_{\theta}(\boldsymbol{x}_{t-1}|\boldsymbol{x}_t))}_{L_{t-1}}
\nonumber\\
&\qquad\qquad
- \underbrace{\log p_{\theta}(\boldsymbol{x}_0|\boldsymbol{x}_1)}_{L_0}
\Bigg].
\end{align}
\vspace{-0.6em}

\noindent In the fourth line, we apply Bayes' rule to rewrite the forward transition as:
\[
q(\boldsymbol{x}_t|\boldsymbol{x}_{t-1}) = \frac{q(\boldsymbol{x}_{t-1}|\boldsymbol{x}_t,\boldsymbol{x}_0)q(\boldsymbol{x}_t|\boldsymbol{x}_0)}{q(\boldsymbol{x}_{t-1}|\boldsymbol{x}_0)}.
\]
This factorization follows from the Markov structure of the forward process. It should not be interpreted as directly replacing $q(\boldsymbol{x}_{t-1}|\boldsymbol{x}_t)$ with $q(\boldsymbol{x}_{t-1}|\boldsymbol{x}_t,\boldsymbol{x}_0)$, since these conditional distributions are generally different. In the expression above, $L_T$ acts as a constant term because the forward process $q(\boldsymbol{x}_T|\boldsymbol{x}_0)$ is independent of the EI network's training, and the reverse process $p_{\theta}(\boldsymbol{x}_T|\boldsymbol{y})$ is modeled as pure noise with a predefined mean and variance. We approximate $L_0$ as $\log q(\boldsymbol{x}_0|\boldsymbol{x}_1)$\footnote{Such an approximation \cite{edge_effect} is widely adopted in DDPM training and helps mitigate edge effects when estimating $L_0$ (see Part 3.3 in \cite{DDPM}).}, which in turn yields the posterior $q(\boldsymbol{x}_1|\boldsymbol{x}_0)q(\boldsymbol{x}_0)/q(\boldsymbol{x}_1)$. Because $L_0$ is also independent of the training parameters, it can be safely ignored\footnote{If the variance is not fixed but instead learned during DDPM training, both $L_T$ and $L_0$ are no longer constant and thus cannot be ignored.}. The remaining term, $L_{t-1}$, represents the KL divergence between two normal distributions. Thus, the upper bound $L_{ub}$ can be rewritten as:
\vspace{-0.6em}
\begin{align}
L_{ub}
&= \sum_{t=2}^{T} \mathbb{E}\left[
\underbrace{
\mathbb{D}_{KL}(q(\boldsymbol{x}_{t-1}|\boldsymbol{x}_t,\boldsymbol{x}_0)
\| p_{\theta}(\boldsymbol{x}_{t-1}|\boldsymbol{x}_t))
}_{L_{t-1}}\right] \notag\\
&= \sum_{t=2}^{T} \mathbb{E}\left[
\underbrace{
\scalebox{0.90}{$\displaystyle
\mathbb{D}_{KL}\!\left(
\mathcal{N}(\boldsymbol{x}_{t-1};
\boldsymbol{\mu}_q,\boldsymbol{\Sigma}_{q}(t))
\| \mathcal{N}(\boldsymbol{x}_{t-1};
\boldsymbol{\mu}_{\theta},\boldsymbol{\Sigma}_{\theta}(t))
\right)$}
}_{L_{t-1}} \right] \notag\\
&\triangleq
\left\| \boldsymbol{\mu}_q(\boldsymbol{x}_t,\boldsymbol{x}_0,t)
- \boldsymbol{\mu}_{\theta}(\boldsymbol{x}_{t},t) \right\|^2_2
\end{align}

In other words, our objective is to optimize $\boldsymbol{\mu}_q(\boldsymbol{x}_t,\boldsymbol{x}_0,t)$ to match $\boldsymbol{\mu}_{\theta}(\boldsymbol{x}_{t},t)$. By using the formulations for $\boldsymbol{\mu}_{\theta}(\boldsymbol{x}_t,t)$ and $\boldsymbol{\mu}_q(\boldsymbol{x}_t,\boldsymbol{x}_0)$ provided in Eq.\eqref{DDPM_preliminary}\footnote{The derivations for $\boldsymbol{\mu}_{\theta}$ apply similarly to $\boldsymbol{\mu}_q$.}, and substituting $\boldsymbol{x}_0$ in $\boldsymbol{\mu}_{\theta}(\boldsymbol{x}_t,t)$ with $f_{\theta}^{(t)}(\boldsymbol{y})$, we obtain:
\begin{equation}
\begin{aligned}
L_{ub}
&= \left\|
\underbrace{\left(
\frac{\sqrt{\bar{\alpha}_{t-1}}\beta_t}{1-\bar{\alpha}_{t}} \boldsymbol{x}_0
+ \frac{\sqrt{\alpha}_{t}(1-\bar{\alpha}_{t-1})}{1-\bar{\alpha}_{t}}
\boldsymbol{x}_t\right)}_{\textit{true mean}}
\right.\\
&\qquad\left.
- \underbrace{\left(
\frac{\sqrt{\bar{\alpha}_{t-1}}\beta_t}{1-\bar{\alpha}_{t}}
f_{\theta}^{(t)}(\boldsymbol{y})
+ \frac{\sqrt{\alpha}_{t}(1-\bar{\alpha}_{t-1})}{1-\bar{\alpha}_{t}}
\boldsymbol{x}_t\right)}_{\textit{estimated mean by EI network}}
\right\|^2_2 \\
&= \left\| \frac{\sqrt{\bar{\alpha}_{t-1}}\beta_t}{1-\bar{\alpha}_{t}} (\boldsymbol{x}_0-f_{\theta}^{(t)}(\boldsymbol{y}))\right\|^2_2 \\
&= \left\|
\frac{\sqrt{\bar{\alpha}_{t-1}}\beta_t}{1-\bar{\alpha}_{t}}
\left(
\frac{\boldsymbol{x}_t-\sqrt{1-\bar{\alpha}_t} \epsilon_0}
{\sqrt{\bar{\alpha}_t}}
-f_{\theta}^{(t)}(\boldsymbol{y})
\right)\right\|^2_2 \\
&= \left\|
\frac{\beta_t}{\sqrt{\alpha_t}(1-\bar{\alpha}_t)}
\left(
\boldsymbol{x}_t-\sqrt{1-\bar{\alpha}_t}\epsilon_0
- \sqrt{\bar{\alpha}_t}f_{\theta}^{(t)}(\boldsymbol{y})
\right) \right\|^2_2
\end{aligned}
\end{equation}

Here, $\boldsymbol{x}_0$ in the third line is replaced by a function of $\boldsymbol{x}_t$ and $\epsilon_0$ using the reparameterization trick: $\boldsymbol{x}_t = \sqrt{\bar{\alpha}_t}\boldsymbol{x}_0 + \sqrt{1-\bar{\alpha}_t} \epsilon_0$, where $\epsilon_0 \sim \mathcal{N}(0,\text{I})$. Inspired by the work of \cite{chen2021equivariant}, we introduce the equivariant constraint $\boldsymbol{L}(\tilde{\boldsymbol{x}}_t^{'}, \tilde{\boldsymbol{x}}_t^{''})$, where $\tilde{\boldsymbol{x}}_t^{'}$ represents the transformed network output and $\tilde{\boldsymbol{x}}_t^{''}$ is the re-estimation of $\tilde{\boldsymbol{x}}_t^{'}$. Following \cite{chen2021equivariant}, we weight the equivariant constraint loss by $\alpha_{EI}$:
\begin{equation}
\begin{aligned}
L^{(t)}
&= \left\|
\frac{\beta_t}{\sqrt{\alpha_t}(1-\bar{\alpha}_t)}
\left(
\boldsymbol{x}_t-\sqrt{1-\bar{\alpha}_t}\epsilon_0
- \sqrt{\bar{\alpha}_t}f_{\theta}^{(t)}(\boldsymbol{y})
\right) \right\|^2_2 \\
&\quad
+ \alpha_{EI}
\left\|\tilde{\boldsymbol{x}}_t^{'} - \tilde{\boldsymbol{x}}_t^{''} \right\|^2_2
\end{aligned}
\end{equation}

For simplicity, we drop the weighting coefficient of the first term, which reduces the loss function to:
\begin{equation} \label{ablation_training_loss}
\begin{aligned}
L_{simple}^{(t)}
&= \left\|
\boldsymbol{x}_t-\sqrt{1-\bar{\alpha}_t}\epsilon_0
- \sqrt{\bar{\alpha}_t}f_{\theta}^{(t)}(\boldsymbol{y})
\right\|^2_2 \\
&\quad
+ \alpha_{EI}
\left\| \tilde{\boldsymbol{x}}_t^{'} - \tilde{\boldsymbol{x}}_t^{''} \right\|^2_2
\end{aligned}
\end{equation}

In our experiments, we found that $L_{simple}^{(t)}$ yields better sample quality than $L^{(t)}$. A similar observation was reported in the original DDPM work \cite{DDPM}, where removing the scaling coefficient improved overall performance.
\vspace{1cm}  % adds 0.5 cm of vertical space

\refstepcounter{section}
\section*{\centering {Supplementary Material: Re-sampling Step under Noisy Conditions}} \label{Appendix_noisy_case}

The detailed derivation of the modified re-estimation step \eqref{re_estimation_noisy_case} and re-sampling step \eqref{modified_output} in the main body are shown here. In noisy HSI inpainting problem, the observation is assumed corrupted with additive white Gaussian noise with noise strength $\sigma_y$: \\
\begin{equation} \label{noisy_inpainting_for_proof}
 \boldsymbol{y} = \rm M \boldsymbol{x} +\boldsymbol{n}, \quad \quad \boldsymbol{n} \sim  \mathcal{N}(0,\sigma_y^2 I)
\end{equation}

We apply the singular value decomposition to the inpainting mask $\rm M$:
\begin{equation} \label{SVD_M}
 {\rm M=\rm \rm U}diag\{ s_1,s_2...s_q\}\rm \rm V^T
\end{equation}

Hence, the pseudo inverse of $\rm M$ can be written as:
\begin{equation} \label{SVD_M_dagger}
 {\rm M^{\dag} =\rm \rm V} diag\{s_1',s_2'...s_q'\}\rm \rm U^T, \quad s_i' = 
\begin{cases}
 1/s_i   & s_i \neq 0 \\
0        & s_i = 0 
\end{cases}
\end{equation}

Now, if we plug in \eqref{noisy_inpainting_for_proof} into the re-estimation step \eqref{re_estimation_noisy_case}, and denote the singular value decomposition of $\Sigma_t$ as: ${\Sigma_t = \rm U} diag \{ \lambda_{t1},\lambda_{t2}...\lambda_{tq} \} \rm V^T = \rm U \Sigma \rm V^T$, it follows that: \\
\begin{equation}
\begin{aligned} \label{reduced}
 \tilde{\boldsymbol{x}}_{0|t}  
 & = f_{\theta}^{(t)}(\boldsymbol{y})
 - \Sigma_t \rm M^{\dag}
 \left(\rm M \textit{f}_{\theta}^{(t)}(\boldsymbol{y})
 -\boldsymbol{y}\right)\\
 & = f_{\theta}^{(t)}(\boldsymbol{y})
 - \Sigma_t \rm M^{\dag}
 \left(\rm M \textit{f}_{\theta}^{(t)}(\boldsymbol{y})
 - (\rm M \boldsymbol{x} +\boldsymbol{n})\right) \\
 & = f_{\theta}^{(t)}(\boldsymbol{y})
 - \Sigma_t \rm M^{\dag}
 \left(\rm M \textit{f}_{\theta}^{(t)}(\boldsymbol{y})
 - \rm M \boldsymbol{x}\right)
 + \Sigma_t \rm M^{\dag} \boldsymbol{n}  \\
 & =
 \underbrace{
 f_{\theta}^{(t)}(\boldsymbol{y})
 - \Sigma_t \rm M^{\dag}
 \left(\rm M \textit{f}_{\theta}^{(t)}(\boldsymbol{y})
 - \rm M \boldsymbol{x}\right)
 }_{\textit{recovers noiseless case}} \\
 &\quad
 + \underbrace{
 \sigma_y \rm U \Sigma \rm V^T \rm M^{\dag}\boldsymbol{\epsilon}
 }_{{\color{red}\textit{introduced noise}}},
 \quad
 \boldsymbol{\epsilon} \sim  \mathcal{N}(0,{\rm I}) 
\end{aligned}
\end{equation}

The noise term is introduced by the observation $\boldsymbol{y}$, which will be further introduced into the DDPM backward pass \eqref{backward_pass} as:
\begin{equation} \label{influenced_output_consider_noise}
\begin{aligned}
\boldsymbol{x}_{t-1} 
&= \frac{\sqrt{\bar{\alpha}_{t-1}}\beta_t}{1-\bar{\alpha}_{t}}
\tilde{\boldsymbol{x}}_{0|t}
+ \frac{\sqrt{\alpha}_{t}(1-\bar{\alpha}_{t-1})}{1-\bar{\alpha}_t}
\boldsymbol{x}_t \\
&\quad + \sigma_t\boldsymbol{\epsilon},
\quad \epsilon \sim  \mathcal{N}(0,{\rm I}) \\
&= \frac{\sqrt{\bar{\alpha}_{t-1}}\beta_t}{1-\bar{\alpha}_{t}}
\tilde{\boldsymbol{x}}_{0|t}
+ \frac{\sqrt{\alpha}_{t}(1-\bar{\alpha}_{t-1})}{1-\bar{\alpha}_{t}}
\boldsymbol{x}_t \\
&\quad
+ \underbrace{
\frac{\sqrt{\bar{\alpha}_{t-1}}\beta_t}{1-\bar{\alpha}_{t}}
\sigma_y \rm U \Sigma \rm V^T \rm M^{\dag} \boldsymbol{\epsilon}
}_{{\color{red}\textit{denoted as $\boldsymbol{\epsilon}_{\mathrm{introduced}}$}}}
+ \sigma_t\boldsymbol{\epsilon}\\
\end{aligned}
\end{equation}

In the above expression, the overall noise level of \eqref{influenced_output_consider_noise} may exceed the pre-defined noise level in  $q(\boldsymbol{x}_{1:T}|\boldsymbol{x}_0)$. In order to ensure the consistency, we construct a new noise term: $\boldsymbol{\epsilon}_{\mathrm{con}}$ in the update of $\boldsymbol{x}_{t-1} $: \\
\begin{equation} \label{modified_output_consider_noise}
\begin{aligned}
\boldsymbol{x}_{t-1} 
\!&= \frac{\sqrt{\bar{\alpha}_{t-1}}\beta_t}{1-\bar{\alpha}_{t}}
\tilde{\boldsymbol{x}}_{0|t}
+ \frac{\sqrt{\alpha}_{t}(1-\bar{\alpha}_{t-1})}{1-\bar{\alpha}_{t}}
\boldsymbol{x}_t \\
&\quad
+ \boldsymbol{\epsilon}_{\mathrm{introduced}}
+ \boldsymbol{\epsilon}_{\mathrm{con}} \\
\boldsymbol{\epsilon}_{\mathrm{introduced}}
+ \boldsymbol{\epsilon}_{\mathrm{con}}
&\sim  \mathcal{N}(0,\sigma_t^2 {\rm I}) 
\end{aligned}
\end{equation}

To design $\boldsymbol{\epsilon}_{\mathrm{con}}$, we start with the introduced noise term in \eqref{influenced_output_consider_noise}, and by spotting that: \\
\begin{equation} \label{epsilon_introduced}
\begin{aligned}
\boldsymbol{\epsilon}_{\mathrm{introduced}}
&= \frac{\sqrt{\bar{\alpha}_{t-1}}\beta_t}{1-\bar{\alpha}_{t}}
\sigma_y \rm U \Sigma \rm V^T \rm M^{\dag}\boldsymbol{\epsilon} \\
&= \frac{\sqrt{\bar{\alpha}_{t-1}}{\beta}_t}{1-\bar{{\alpha}}_{t}}
\sigma_y {\rm U}{\Sigma} {\rm V^T} \\
&\quad \times
(\rm {\rm V}
\underbrace{diag\{s_1',s_2'...s_q'\}}_{\textit{denoted as ${\Omega}$}}
\rm {\rm U}^T) \boldsymbol{\epsilon}
\end{aligned}
\end{equation}

\noindent which can be re-written as: 
\begin{equation} \label{variance_introduced}
\begin{aligned}
\boldsymbol{\epsilon}_{\mathrm{introduced}}
&\sim  \mathcal{N} \left(
0,
\left(\frac{\sqrt{\bar{{\alpha}}_{t-1}}{\beta}_t}
{1-\bar{{\alpha}}_{t}}\right)^2
\sigma_y^2 \right.\\
&\qquad\qquad\left.
\times ({\rm U}{\Sigma} {\rm V^T}{\rm M}^{\dag})
(({\rm M}^{\dag})^T {\rm V}{\Sigma} {\rm U}^T )
\right)\\
&\sim  \mathcal{N} \left(
0,
\left(\frac{\sqrt{\bar{{\alpha}}_{t-1}}{\beta}_t}
{1-\bar{{\alpha}}_{t}}\right)^2
\sigma_y^2 \right.\\
&\qquad\qquad\left.
\times ({\rm U}{\Sigma} {\rm V^T}{\rm V} {\Omega}{\rm U}^T)
({\rm U}{\Omega}{\rm V}^T {\rm V}{\Sigma} {\rm U}^T )
\right)\\
&\sim  \mathcal{N} \left(
0,
\left(\frac{\sqrt{\bar{{\alpha}}_{t-1}}{\beta}_t}
{1-\bar{{\alpha}}_{t}}\right)^2
\sigma_y^2
({\rm U}{\Sigma} {\Omega}{\Omega}{\Sigma} {\rm U}^T )
\right)\\
\end{aligned}
\end{equation} 

Recall the definition of ${\Sigma}$ and ${\Omega}$ are:
\begin{equation} \label{Omega_and_Sigma}
\begin{aligned}
{\Sigma} &= diag\{{\lambda_{t1}},{\lambda_{t2}}...{\lambda_{tq}\}}, \\
{\Omega} &= diag\{s_1',s_2'...s_q'\}, \quad s_i' = 
\begin{cases}
 1/s_i   & s_i \neq 0 \\
0        & s_i = 0 
\end{cases}
\end{aligned}
\end{equation}

Plug in \eqref{Omega_and_Sigma} into \eqref{variance_introduced} yields:
\begin{equation} \label{compact_form_variance_introduced}
\begin{aligned}
\boldsymbol{\epsilon}_{\mathrm{introduced}}
&\sim  \mathcal{N} (0,{\rm U} {\rm \Lambda_t} {\rm U}^T ), \\
{\Lambda_{ti}}  &= 
\begin{cases}
\frac{
\left(\frac{\sqrt{\bar{{\alpha}}_{t-1}}{\beta}_t}
{1-\bar{{\alpha}}_{t}}\right)^2
\sigma_y^2 {\lambda_{ti}}^2}{{s_i}^2}
& s_i \neq 0 \\
0        & s_i = 0 
\end{cases} 
\end{aligned}
\end{equation} 

Note that $\boldsymbol{\epsilon}_{\mathrm{introduced}}$ is a function of the noise level $\sigma_y$ of the observation $\boldsymbol{y}$ (assumed known) and the scaling matrix $\Sigma_t$. $\Sigma_t$ is designed to guarantee that the noise variance at each update step remains bounded by $\sigma_t$ (noise scheduling in the forward/diffusion process). Hence, we define $\boldsymbol{\epsilon}_{\mathrm{con}}$ as: \\
\begin{equation}
\begin{aligned}
\boldsymbol{\epsilon}_{\mathrm{con}}
&\sim \mathcal{N}(\boldsymbol{0},\boldsymbol{\Gamma}_t),\\
\boldsymbol{\Gamma}_t &= \operatorname{diag}\{\Gamma_{t1},\Gamma_{t2},\ldots,\Gamma_{tq}\},\\
\Gamma_{ti} &= 
\begin{cases}
\sigma_t^2 -
\frac{
\left(\frac{\sqrt{\bar{\alpha}_{t-1}}\beta_t}{1-\bar{\alpha}_{t}}\right)^2
\sigma_y^2 {\lambda_{ti}}^2}{s_i^2}
& s_i \neq 0 \\
 \sigma_t^2  & s_i = 0 
\end{cases}
\end{aligned}
\end{equation}

The choice of $\Sigma_t$ should satisfy that: (1) When the noise level of the current estimate is relatively big, it should be set as close to $\rm {I}$ as possible to better utilize the information in the observation $\boldsymbol{y}$. (2) Otherwise, it should be scaled accordingly. Following the design of DDNM \cite{DDNM}, we choose $\lambda_{ti}$ using the following thresholding rule to limit the contribution of the observation noise $\sigma_y$ and keep the total noise level consistent with the diffusion schedule: \\
\begin{equation} \label{Sigma_t}
 \lambda_{ti} =
\begin{cases}
1
& \sigma_t \ge
\frac{
\left(\frac{\sqrt{\bar{\alpha}_{t-1}}\beta_t}{1-\bar{\alpha}_{t}}\right)
\sigma_y}{s_i} \\
\frac{\sigma_ts_i}
{\left(\frac{\sqrt{\bar{\alpha}_{t-1}}\beta_t}{1-\bar{\alpha}_{t}}\right)
\sigma_y}
& \sigma_t <
\frac{
\left(\frac{\sqrt{\bar{\alpha}_{t-1}}\beta_t}{1-\bar{\alpha}_{t}}\right)
\sigma_y}{s_i} \\
1  & s_i = 0 \\
\end{cases}
\end{equation}

\vspace{1cm}  % adds 0.5 cm of vertical space
\refstepcounter{section}
\section*{\centering {Supplementary Material: Offline Training Capability}} \label{Offline_training}

While the standard HyDiff-EI framework presented in the main text performs per-image optimization at test time (in a DIP-like manner), our approach is not fundamentally restricted to single-image optimization. It can be readily adapted to pre-train on a dataset of corrupted HSI measurements, functioning as an unsupervised, measurement-only trained model. This flexibility offers superiority over existing DIP-based methods, which generally do not provide pre-training or fine-tuning capabilities. To demonstrate in more detailed, we trained the EI network within our proposed diffusion solver using 300 corrupted HSI samples from the Chikusei dataset. The images were randomly corrupted using the mask types shown in Figure \ref{HSI_inpainging_noiseless} of the main text.
\begin{figure}[h]
    \centering
    \includegraphics[width=\linewidth]{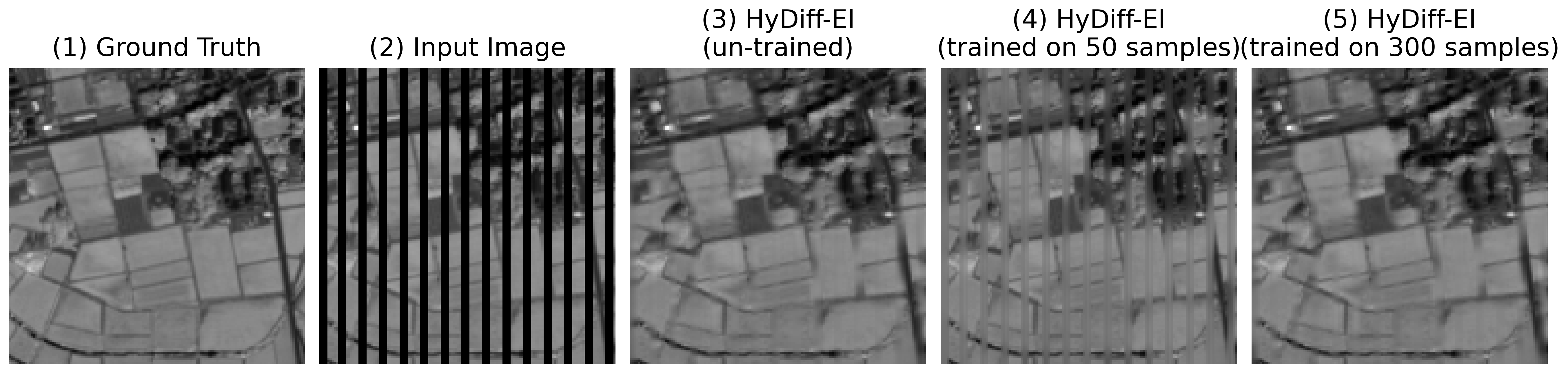}
    \caption{Visual comparison of inpainting results produced by the trained and un-trained HyDiff-EI models.}
    \label{figure_Offline_training_performance}
\end{figure}

During training, we used a batch size of 5 and the Adam optimizer with a learning rate of 0.0001. The network was trained for 20000 epochs. In each epoch, for every mini-batch, we uniformly sampled a single random diffusion timestep $t \sim \mathcal{U}(0, 1000)$, computed the training loss (Eq. \ref{ablation_training_loss}) at that specific step, and performed a single parameter update on $f_{\theta}^{(t)}(\boldsymbol{y})$. Following this offline training phase, the pre-trained EI network $f_{\theta}^{(t)}(\boldsymbol{y})$ was integrated back into the diffusion framework for inpainting. We refer to this trained model as HyDiff-EI (trained) to distinguish it from the original, test-time optimized HyDiff-EI (untrained) presented earlier in the main text.

Table \ref{table_Offline_training_run_time} presents the quantitative inpainting results for the un-trained HyDiff-EI model alongside variants trained on 50 and 300 corrupted HSI samples, while Figure \ref{figure_Offline_training_performance} illustrates their visual performance. 

\begin{table}[h]
\centering
\caption{Inpainting performance of the un-trained HyDiff-EI model and two variants trained on 50 and 300 corrupted HSI samples, respectively, evaluated on a $144 \times 144 \times 128$ Chikusei sample.} 
\scalebox{0.8}{
\begin{tabular}{lcccccc}
\toprule
\textbf{Method} & Input & Un-trained & Trained (50 samples) & Trained (300 samples) \\
\midrule
\textbf{Time (s/image)} & \diagbox[]{}{} & 38.02 & 15.15 & 15.15  \\
\textbf{MPSNR (dB)}$\uparrow$ & 20.704 & 34.393 & 31.113 &34.540 \\
\textbf{MSSIM}$\uparrow$ & 0.186 & 0.927 & 0.863 & 0.929  \\ 
\bottomrule
\end{tabular}
}
\label{table_Offline_training_run_time}
\end{table} 
It can be seen that training on prior measurement data effectively reduces the optimization time during inference. The remaining 15.15 seconds is expected, as the model still needs to go through the whole diffusion process (1000 diffusion steps in this example). These results demonstrate that the proposed HyDiff-EI framework is not restricted solely to test-time optimization. Given sufficient (corrupted) training measurements, the model yields satisfactory results at test time. This enhanced flexibility makes it highly appealing for real-world deployment.

\end{document}